\documentclass[10pt,twocolumn,letterpaper]{article}

\usepackage{cvpr}              % To produce the CAMERA-READY version
\usepackage[accsupp]{axessibility}  % Improves PDF readability for those with disabilities.

\definecolor{cvprblue}{rgb}{0.21,0.49,0.74}
\usepackage[pagebackref,breaklinks,colorlinks,allcolors=cvprblue]{hyperref}
\usepackage{multirow}
\usepackage{pifont}
\usepackage{colortbl}
\usepackage{graphicx}
\usepackage{subcaption} % 导入subcaption包
\title{IDEA: Inverted Text with Cooperative Deformable Aggregation for Multi-modal Object Re-Identification}
\author{
    Yuhao Wang,
    Yongfeng Lv,
    Pingping Zhang\thanks{Corresponding author},
    Huchuan Lu\\
    School of Future Technology, School of Artificial Intelligence, Dalian University of Technology \\
    {\tt\small \{924973292,lvyf\}@mail.dlut.edu.cn, \{zhpp,lhchuan\}@dlut.edu.cn}
}

\begin{document}
\maketitle
\begin{abstract}
    Multi-modal object Re-IDentification (ReID) aims to retrieve specific objects by utilizing complementary information from various modalities.
    However, existing methods focus on fusing heterogeneous visual features, neglecting the potential benefits of text-based semantic information. 
    To address this issue, we first construct three text-enhanced multi-modal object ReID benchmarks.
    To be specific, we propose a standardized multi-modal caption generation pipeline for structured and concise text annotations with Multi-modal Large Language Models (MLLMs).
    Besides, current methods often directly aggregate multi-modal information without selecting representative local features, leading to redundancy and high complexity.
    To address the above issues, we introduce IDEA, a novel feature learning framework comprising the Inverted Multi-modal Feature Extractor (IMFE) and Cooperative Deformable Aggregation (CDA).
    The IMFE utilizes Modal Prefixes and an InverseNet to integrate multi-modal information with semantic guidance from inverted text.
    The CDA adaptively generates sampling positions, enabling the model to focus on the interplay between global features and discriminative local features.
    With the constructed benchmarks and the proposed modules, our framework can generate more robust multi-modal features under complex scenarios.
    Extensive experiments on three multi-modal object ReID benchmarks demonstrate the effectiveness of our proposed method.
\end{abstract}
\section{Introduction}
\label{sec:intro}
Object Re-IDentification (ReID) focuses on retrieving the same object across different camera views. 
While significant progress has been made with RGB images~\cite{he2021transreid,liu2021watching,zhang2021hat,wang2021pyramid,liu2023deeply,shi2024learning,wang2024other,liu2024video,yu2024tf,yang2024shallow,gong2024cross,wang2025unity}, existing methods are limited under environmental challenges like dark and glare, reducing their robustness in real-world applications.
Multi-modal imaging, which combines data from multiple spectra such as RGB, Near Infrared (NIR) and Thermal Infrared (TIR), offers a promising solution. 
By leveraging complementary information, multi-modal ReID methods~\cite{lu2023learning,crawford2023unicat,wang2024top,zhang2024magic} improve feature robustness under challenging conditions.

Beyond visual information fusion for different spectra, integrating language with vision can enhance ReID performance~\cite{han2024clip,wang2024large,niu2025chatreid}.
However, due to the absence of text annotations for images, most existing methods rely solely on visual features, overlooking the benefits of text-based semantic information.
To address this limitation, some studies augment RGB datasets with manual text annotations~\cite{li2017person,ding2021semantically}.
While this strategy boosts performance, it is both time-consuming and labor-intensive.
\begin{figure}[t]
  \centering
    \resizebox{0.475\textwidth}{!}
    {
  \includegraphics[width=30\linewidth]{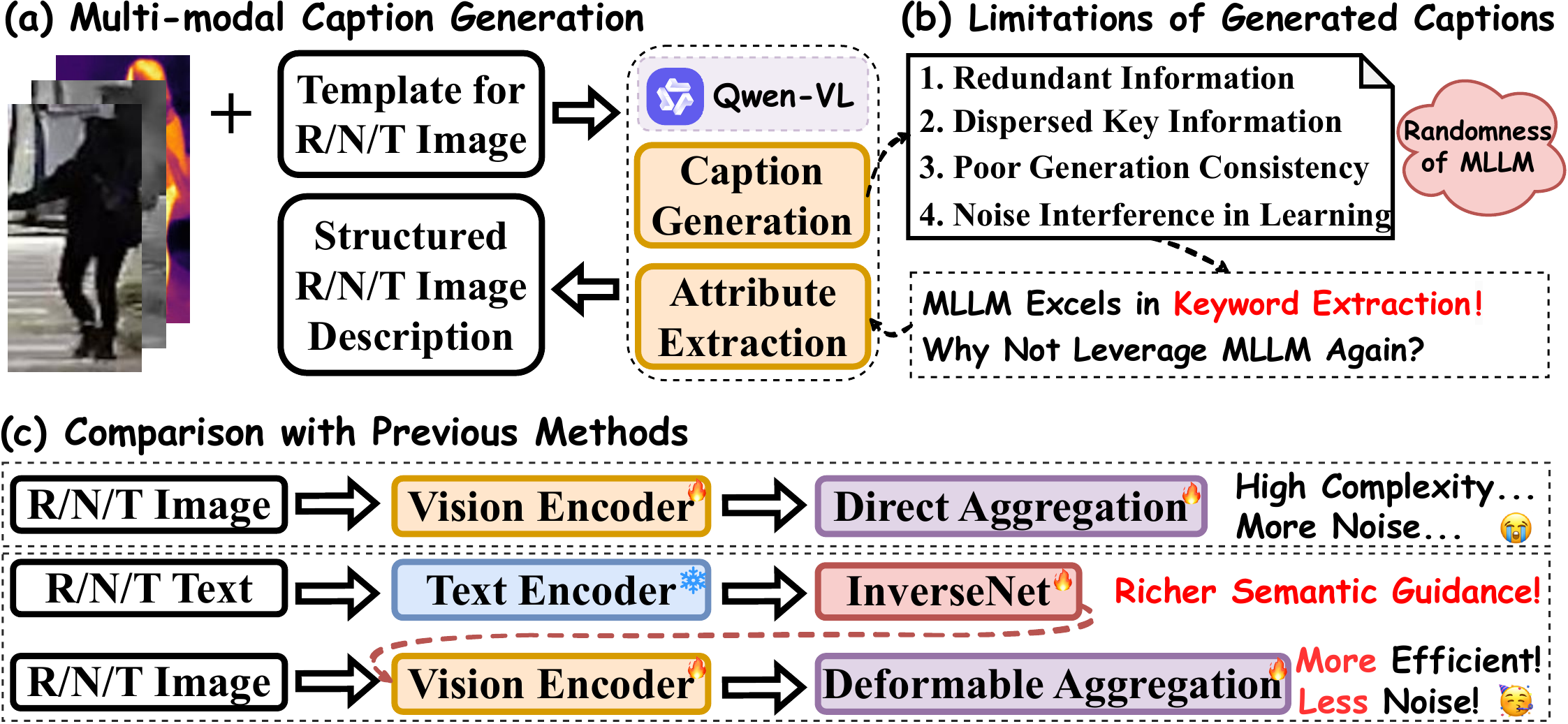}
  }
  \vspace{-6mm}
   \caption{Overall illustration of our motivations and proposed framework.
   (a) Our multi-modal caption generation pipeline.
   (b) Limitations of existing MLLM-generated captions.
   (c) Comparison between previous methods and our proposed IDEA.
   }
  \label{fig:LLM}
  \vspace{-3mm}
\end{figure}
%～～～～～～～～～～～～～～～～～～～～～～～～～～～～～～～～
With the advent of Multi-modal Large Language Models (MLLMs), image caption has made significant progress. 
Recently, researchers start to use MLLMs to generate textual descriptions for RGB images~\cite{han2024clip,he2024instruct}.
However, these methods face two main challenges.
(1) As shown in \textcolor{red}{Fig.}~\ref{fig:LLM} (b), the randomness in caption generation often leads to redundancy, resulting in long and complex sentences. 
Additionally, the text structure varies across images.
In ReID tasks, such captions may exceed input length limits of a text encoder~\cite{radford2021learning,zhang2024multimodal}, with redundant content diluting the essential information.
(2) Current methods primarily focus on RGB images, where the generated captions lack sufficient details.
However, multi-modal imaging can capture crucial information in complex environments, providing enough information for MLLMs to generate informative captions.
\textbf{Despite this, no existing methods provide text annotations for multi-modal images.} 
To bridge this gap, we construct three text-enhanced multi-modal object ReID benchmarks.
Meanwhile, another challenge is the aggregation of multi-modal information.
As shown in \textcolor{red}{Fig.}~\ref{fig:LLM} (c), previous methods directly aggregate the heterogeneous information, leading to high computational complexity and noise interference~\cite{zhang2024magic}.
To address above issues, we propose a multi-modal feature learning framework named IDEA to introduce \textbf{I}nverted text with cooperative \textbf{DE}formable \textbf{A}ggregation for multi-modal object ReID.

Technically, we first develop a standardized pipeline for multi-modal caption generation.
It extends existing datasets with text annotations across spectral modalities. 
Specifically, as shown in \textcolor{red}{Fig.}~\ref{fig:LLM} (a), our caption generation pipeline consists of two steps.
In the first step, we use image-template pairs in conjunction with MLLMs to generate informative captions.  
Then, we leverage the powerful keyword extraction capabilities of MLLMs to extract predefined attributes from the generated captions to produce structured and concise sentences.
Building upon the captions, we propose the IDEA framework, which consists of two key components: the Inverted Multi-modal Feature Extractor (IMFE) and Cooperative Deformable Aggregation (CDA). 
First, we utilize the IMFE to extract integrated multi-modal features using Modal Prefixes and an InverseNet. 
Specifically, Modal Prefixes are designed to distinguish different modalities, reducing the impact of conflicting information across modalities. 
The InverseNet further exploits the semantic information from inverted text to enhance feature robustness.
Furthermore, we propose CDA to adaptively aggregate discriminative local information. 
Specifically, based on the aggregated multi-modal information, we adaptively generate sampling positions to enable the model to focus on the interplay between global features and discriminative local information.
Through these modules, our proposed framework effectively utilizes semantic guidance from texts while adaptively aggregating discriminative multi-modal information.
Extensive experiments on three benchmark datasets demonstrate the effectiveness of our approach.
In summary, our contributions are as follows:
\begin{itemize}
  \item We construct three text-enhanced multi-modal object ReID benchmarks, providing a structured caption generation pipeline across multiple spectral modalities.
  \item We introduce a novel feature learning framework named IDEA, which includes the Inverted Multi-modal Feature Extractor (IMFE) and Cooperative Deformable Aggregation (CDA). 
  The IMFE integrates multi-modal features using Modal Prefixes and an InverseNet, while the CDA adaptively aggregates discriminative local information.
  \item Extensive experiments on three benchmark datasets validate the effectiveness of our proposed method.
\end{itemize}
\section{Related Work}
\label{sec:related}
\begin{figure*}[t]
    \centering
      \resizebox{1.0\textwidth}{!}
      {
    \includegraphics[width=30\linewidth]{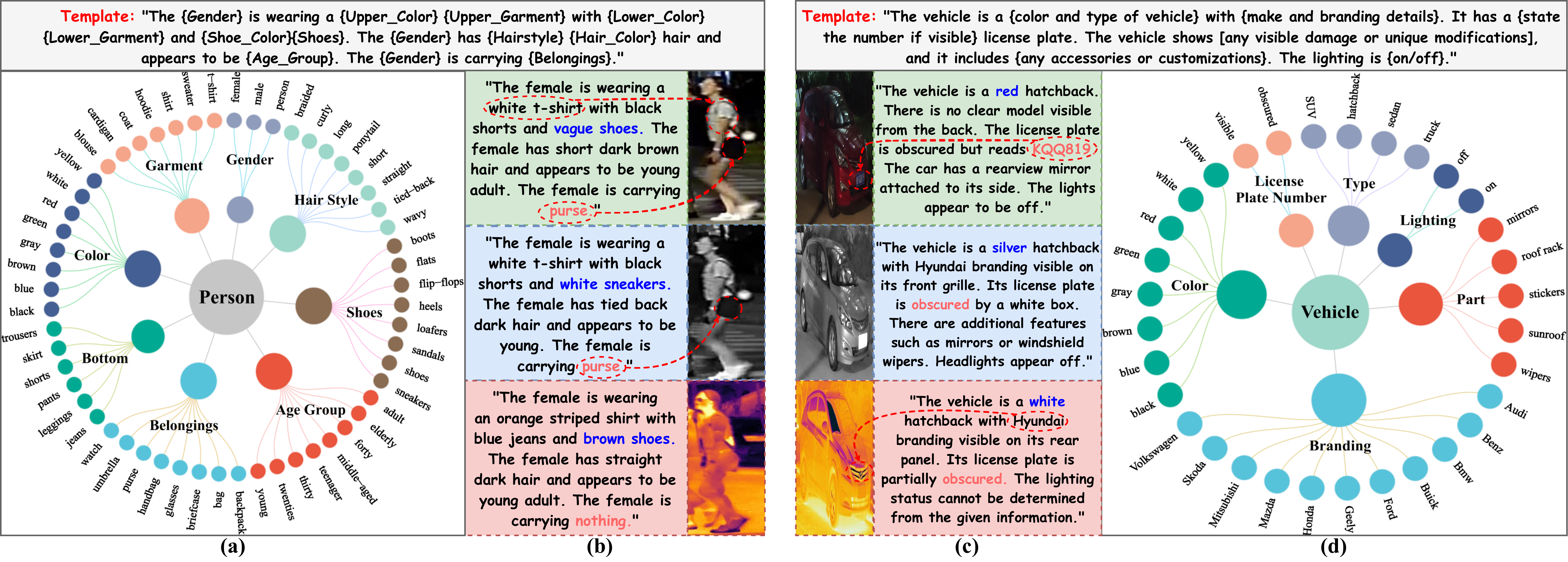}
    }
    \vspace{-6mm}
     \caption{The upper row presents the template used to annotate our dataset.
     The lower row provides detailed information about the annotated dataset.
     (a) Category statistics for our annotated person ReID dataset.
     (b) Example images and captions from the RGBNT201 dataset.
     (c) Category statistics for our annotated vehicle ReID dataset.
     (d) Example images and captions from the MSVR310 dataset.}
    \label{fig:Dataset}
    \vspace{-2mm}
  \end{figure*}
%～～～～～～～～～～～～～～～～～～～～～～～～～～～～～～～～
\subsection{Vision-language Learning in Re-Identification}
Vision-language learning plays a pivotal role in multi-modal applications.
As the field advances~\cite{yu2024boosting,diao2024unveiling,yu2024llms}, there is a growing focus on exploring the interaction between visual and textual information in ReID tasks.
Existing methods can be broadly divided into cross-modal and multi-modal ReID. 
In cross-modal ReID, Text-to-Image ReID~\cite{ding2021semantically} focuses on matching text queries with image galleries. 
Current methods primarily emphasize feature alignment~\cite{jiang2023cross} and pre-training~\cite{tan2024harnessing}. 
To expand application scenarios, Zhai et al.~\cite{zhai2022trireid} combine text and sketch modalities in the query, while Chen et al.~\cite{chen2023towards} propose an uncertain Query-to-RGB retrieval model to handle missing modalities. 
More recently, Li et al.~\cite{li2024all} introduce flexible modality combinations and unify various uncertainties in retrieval.
However, these methods focus on cross-modal alignment.
In contrast, multi-modal ReID aims to leverage complementary information from multiple modalities, with both the query and gallery containing multi-modal data. 
The rise of vision-language models, particularly CLIP~\cite{radford2021learning}, has significantly advanced this area by facilitating image-text interactions. 
Li et al.~\cite{li2023clipreid} pioneer CLIP-ReID to learn from text templates for ReID tasks. 
However, the templates are not real descriptions of images, potentially restricting performance. 
To address this, Han et al.~\cite{han2024clip} leverage MLLMs to generate descriptions for RGB images.
Despite improvements, existing methods fail to tackle the challenges of generating text for multi-spectral modalities and ensuring structural consistency in the generated descriptions. 
To address the above issues, we propose a structured multi-modal caption generation pipeline, enhancing the consistency of texts and providing informative annotations for multi-modal object ReID.
\subsection{Multi-modal Object Re-Identification}
Multi-modal object ReID gains great attention due to its robustness in real-world applications. 
Research primarily focuses on leveraging complementary information from different modalities. 
Early works emphasize effective fusion strategies~\cite{gong2021eliminate,zheng2021robust}. 
To enhance modality-specific learning, Wang et al.~\cite{wang2022interact} propose an interact-embed-enlarge framework to facilitate knowledge expansion. 
To address missing modal data, Zheng et al.~\cite{zheng2023dynamic} introduce pixel reconstruction to handle data inconsistency. 
Li et al.~\cite{li2020multi} use a coherence loss to guide feature fusion. 
Some methods further improve feature robustness with modality generation, graph learning and instance sampling~\cite{he2023graph, guo2022generative, zheng2023cross}. 
%
% The advent of vision Transformers (ViTs)~\cite{dosovitskiy2020image} shifts research toward Transformer-based architectures due to their superior generalization capabilities. 
%
%Pan et al.~\cite{pan2023progressively} and Crawford et al.~\cite{crawford2023unicat} utilize attention mechanisms to model complex cross-modal relationships.
%
Notably, Wang et al.~\cite{wang2024heterogeneous} introduce a test-time training framework to mine inter-modal interactions. 
Wang et al.~\cite{wang2024top} develop TOP-ReID, which employs token permutation to guide modality fusion.
Recently, Zhang et al.~\cite{zhang2024magic} propose diverse token selections to mitigate background noises.
Despite promising results, they often overlook the semantic guidance of informative text annotations.
Thus, we propose the IDEA framework, which incorporates generated text annotations to leverage semantic guidance.
This framework adaptively aggregates discriminative local information, enhancing feature robustness in complex scenarios.
\section{Proposed Method}
\label{sec:methods}
\begin{figure*}[t]
    \centering
      \resizebox{1.0\textwidth}{!}
      {
    \includegraphics[width=30\linewidth]{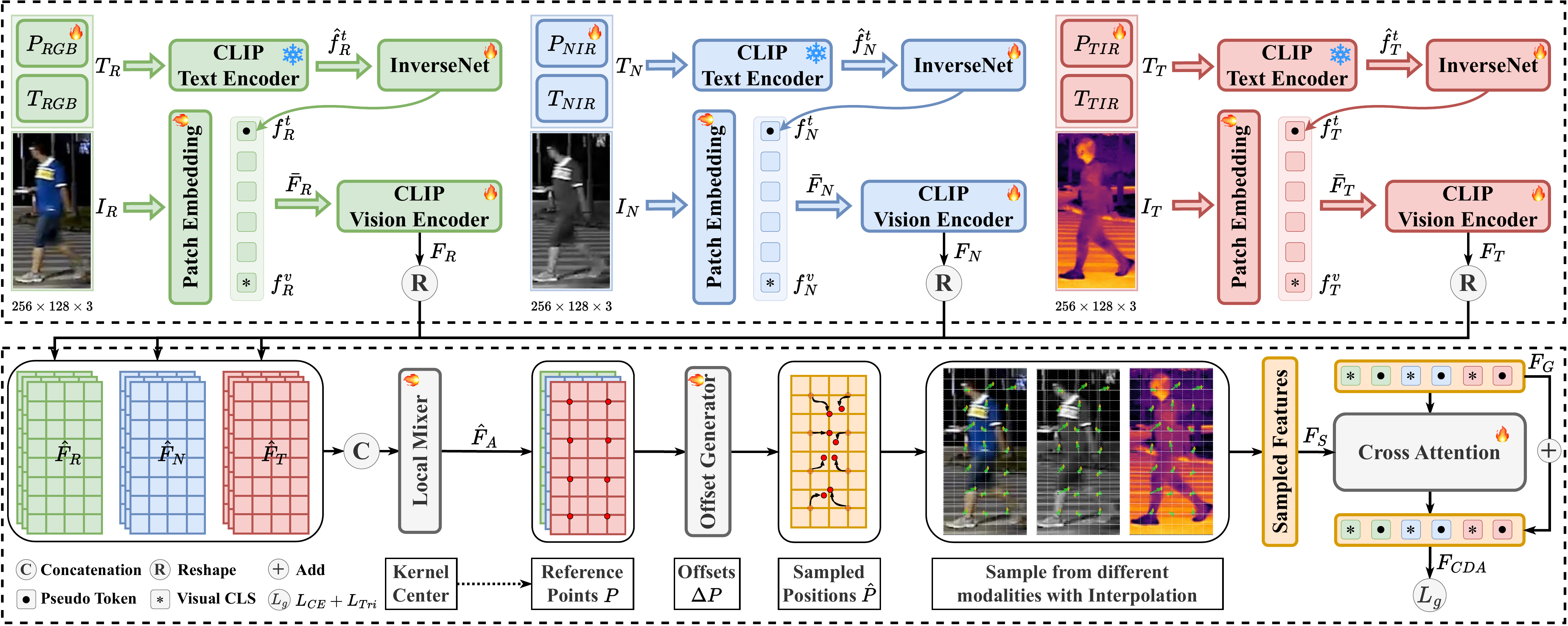}
    }
    \vspace{-4mm}
     \caption{Illustration of the proposed IDEA framework.
     The upper part depicts the Inverted Multi-modal Feature Extractor (IMFE).
     It employs modal prefixes and an InverseNet to incorporate semantic text guidance for feature discriminability.
     The lower part highlights the Cooperative Deformable Aggregation (CDA), which adaptively integrates discriminative local information with global features.
     With the integration of IMFE and CDA, IDEA effectively extracts discriminative multi-modal features for object ReID.}
    \label{fig:Overall}
    \vspace{-2mm}
  \end{figure*}

In this section, we introduce the multi-modal caption generation pipeline and the proposed feature learning framework.
\textcolor{red}{Fig.}~\ref{fig:Dataset} illustrates the details of caption generation, including attribute definitions and examples.
\textcolor{red}{Fig.}~\ref{fig:Overall} presents the key modules of our proposed IDEA: the Inverted Multi-modal Feature Extractor (IMFE) and the Cooperative Deformable Aggregation (CDA).
Details are described as follows.
  %～～～～～～～～～～～～～～～～～～～～～～～～～～～～～～～～
\subsection{Multi-modal Caption Generation}
To bridge the gap in multi-modal text annotation, we propose the Multi-modal Caption Generation pipeline.
To be specific, our pipeline consists of two steps.
The first step generates informative descriptions, while the second step reuses the MLLMs to extract the predefined key attributes.
\\
\textbf{Caption Generation.}
Taking our person annotation as an example, we define 8 attribute categories in \textcolor{red}{Fig.} \ref{fig:Dataset} (a), including gender, belongings and so on.
These attributes are mapped into the template shown in the top part of \textcolor{red}{Fig.} \ref{fig:Dataset} (a).
To tailor the description to each modality, we add specific prefixes: \textit{“Write a comprehensive description of the person's overall appearance based on the [RGB/NIR/TIR] image, ...”}.
Besides, we incorporate commands to prevent the hallucination of the MLLMs~\cite{tan2024harnessing}, forming a complete prompt.
This prompt, along with the corresponding image, is fed into the MLLMs to generate the textual description.
\\
\textbf{Attribute Extraction.}
After obtaining an initial description, we feed it back into the MLLMs to extract predefined attributes.
After that, we populate these attributes into the same template we used for caption generation, forming a structured and concise description.
As shown in \textcolor{red}{Fig.}~\ref{fig:Dataset} (b), the descriptions for each modality can align well with the corresponding images.
For instance, the \textit{“white t-shirt”} and \textit{“purse”} in the RGB image are accurately captured.
For the vehicle dataset, we follow similar pipelines.
As illustrated in \textcolor{red}{Fig.}~\ref{fig:Dataset} (c), the text of the RGB image successfully captures the license plate number \textit{“KQQ819”}.
Meanwhile, the car logo is more prominent in the TIR image, with \textit{“Hyundai”} being identified.
This approach allows us to establish a unified pipeline for multi-modal text annotation, ensuring structural descriptions across different modalities.
\subsection{IDEA Framework}
\subsubsection{Inverted Multi-modal Feature Extractor}
To leverage the semantic guidance from texts, we propose the Inverted Multi-modal Feature Extractor (IMFE).
Unlike previous methods~\cite{wang2024top,zhang2024magic} that focus on fusing multi-modal images, we incorporate multi-modal texts into the fusion.
However, as indicated by the \textcolor{blue}{blue keywords} in \textcolor{red}{Fig.}~\ref{fig:Dataset} (b) and (c), conflicts can arise due to the nature of multi-modal images.
Directly aggregating contradictory information can lead to model confusion.
Thus, IMFE employs Modal Prefixes and an InverseNet to address these issues.\\
\textbf{Modal Prefixes.}
To enhance the model’s awareness of different modal texts and mitigate the impact of conflicting information, we propose a simple method called Modal Prefixes.
Taking the text input for the RGB branch as an example, as shown in the top left corner of \textcolor{red}{Fig.}~\ref{fig:Overall}, $T_{RGB}$ represents the text annotation of the RGB image, while $P_{RGB}$ is the modal prefix for RGB modality.
$P_{RGB}$ consists of two parts: a fixed text describing the characteristics of the RGB image and learnable tokens for fine-tuning.
Specifically, $P_{RGB}$ is structured as: \textit{“An image of a XXXX person in the visible spectrum, capturing natural colors and fine details: ”}, where \textit{“XXXX”} is replaced by an equal number of learnable tokens~\cite{li2023clip}.
Here, we denote the number of learnable tokens as $N_{P}$.
Then, we concatenate $P_{RGB}$ and $T_{RGB}$ to form the text input $T_{R}$ for the RGB branch:
\begin{equation}
    T_{R} = [P_{RGB}, T_{RGB}].
\end{equation}
Other modalities follow the same structure to guide the model in distinguishing between different modal texts.
With this simple yet effective method, the model can better understand the characteristics of each modality and reduce conflicts during multi-modal fusion.
\\
\textbf{InverseNet.}
To fully leverage the semantic information from texts, we propose the InverseNet.
Previous methods~\cite{baldrati2023zero,han2024clip} often reverse global image information into pseudo-text tokens and use texts as the primary modality for downstream tasks.
However, due to the limitations of MLLMs and the complexity of multi-modal image annotations, the generated texts may contain some errors and conflicts.
To address this, we take a different direction, reversing global text information into pseudo-image tokens.
In this reversal, we also incorporate rich information from the learnable tokens in Modal Prefixes.
This approach enables the interaction between semantic information and image details, resulting in a discriminative feature representation.

Specifically, as shown in the upper part of \textcolor{red}{Fig.}~\ref{fig:Overall}, the text $T_{m}$ is first fed into the CLIP's text encoder $\mathcal{T}$ for extracting the semantic feature $\hat{f}^{t}_{m} \in \mathbb{R}^{C}$ as follows:
\begin{equation}
    \hat{f}^{t}_{m} = \mathcal{T}(T_{m}).
\end{equation}
Here, $m \in \{R, N, T\}$ represents the RGB, NIR and TIR modalities, respectively.
$C$ denotes the embedding dimension.
If learnable tokens exist in the modal prefixes, $\hat{f}^{t}_{m}$ will be updated by the average of learnable tokens.
Then, we feed $\hat{f}^{t}_{m}$ into the InverseNet $\mathcal{I}$ to generate the pseudo token $f^{t}_{m} \in \mathbb{R}^{C'}$ with the following equation:
\begin{equation}
    f^{t}_{m} = \mathcal{I}(\hat{f}^{t}_{m}).
\end{equation}
Here, $C'$ is the dimension of the pseudo token \textbf{while $\mathcal{I}$ is essentially a simple layer of MLP}.
Meanwhile, for the image input $I_{m}$, we first patchify it into $N_l$ patches.
Then, we concatenate the pseudo token \( f^{t}_{m} \), the patch tokens and a learnable token \( f^{v}_{m} \) to form the image feature \( \bar{F}_{m} \in \mathbb{R}^{(N_l + 2)\times C'} \).
Finally, we feed \( \bar{F}_{m} \) into the CLIP's vision encoder \( \mathcal{V} \) to obtain the integrated feature \( F_{m} \in \mathbb{R}^{(N_l + 2) \times C} \) as follows:
\begin{equation}
    F_{m} = \mathcal{V}(\bar{F}_{m}).
\end{equation}
By incorporating the InverseNet, we can effectively leverage the semantic information from texts and enhance the discriminability of the feature representation.

\subsubsection{Cooperative Deformable Aggregation}
To effectively extract discriminative local information and adaptively aggregate complementary multi-modal features, we propose Cooperative Deformable Aggregation (CDA).
%
% Existing methods~\cite{wang2022interact,wang2024top} often involve interactions using all patch tokens across modalities, leading to two challenges: high computational cost and vulnerability to noise in local patches, which can undermine feature robustness during multi-modal fusion.
%
CDA addresses the noise interference by adaptively generating sampling positions of key local regions.
This module enables interactions between global features and discriminative local features, ensuring efficient feature aggregation.

Technically, as illustrated in the lower part of \textcolor{red}{Fig.}~\ref{fig:Overall}, we first extract patch tokens from the integrated feature \( F_{m} \).
These patches are then reshaped to form \( \hat{F}_{m} \in \mathbb{R}^{H \times W \times C} \), where \( H \) and \( W \) represent the height and width of the local feature map, respectively.
Next, we concatenate these features along the channel dimension and pass them into the Local Mixer \( \mathcal{M} \) for interaction.
Specifically, \( \mathcal{M} \) comprises point-wise convolution \( \mathcal{P} \), the GELU activation function~\cite{hendrycks2016gaussian} \( \delta \) and depth-wise convolution \( \mathcal{D} \).
This process yields the aggregated local features \( \hat{F}_{A} \in \mathbb{R}^{H_{S} \times W_{S} \times C} \):
\begin{equation}
    \hat{F}_{A} = \mathcal{M}([\hat{F}_{R}, \hat{F}_{N}, \hat{F}_{T}]),
\end{equation}
\begin{equation}
  \mathcal{M}(\mathcal{X}) = \delta(\mathcal{D}(\delta(\mathcal{P}(\mathcal{X})))),
\end{equation}
where \([\cdot]\) means concatenation.
\( H_{S} \) and \( W_{S} \) represent the number of sampled regions in the height and width dimensions, respectively.
Since the convolution \( \mathcal{D} \) aggregates local information in the receptive field of the kernel, we define the center of each convolution operation as the reference points \( P \in \mathbb{R}^{H_{S} \times W_{S} \times 2} \).
Based on these reference points, the aggregated local information is then used to determine the direction in which the model should shift within the local region.
Thus, we send \( \hat{F}_{A} \) into the Offset Generator \( \mathcal{G} \) to predict the offsets \( \Delta{P} \in \mathbb{R}^{H_{S} \times W_{S} \times 2} \) as follows:
\begin{equation}
    \Delta{P} = \mathcal{G}(\hat{F}_{A}) * k,
\end{equation}
where \( k \) scales the offset magnitude~\cite{xia2022vision} and \( \mathcal{G} \) denotes a linear layer.
Finally, we apply the offsets \( \Delta{P} \) to the reference points \( P \) to get the sampling locations \( \hat{P} \in \mathbb{R}^{H_{S} \times W_{S} \times 2} \), which are \textbf{shared across modalities}~\cite{zhang2024magic}:
\begin{equation}
    \hat{P} = P + \Delta{P}.
\end{equation}
Next, we sample the local features \( \hat{F}_{m} \) at the positions \( \hat{P} \) using the bilinear interpolation to obtain the discriminative local features.
The features are concatenated to form the sampled feature \( F_{S} \in \mathbb{R}^{3N_{S} \times C} \), where \( N_{S} = H_{S} \times W_{S} \) represents the number of sampled regions.
At this stage, we utilize the cooperative information from multiple modalities to generate deformable local discriminative features.
As illustrated by the sampling points visualization in \textcolor{red}{Fig.}~\ref{fig:Overall}, the adaptive offsets effectively concentrate on critical semantic regions of the human body, enhancing the discrimination.

To promote the interaction between global features and discriminative local features, we incorporate cross attention~\cite{dosovitskiy2020image}.
Specifically, we extract the visual class token and pseudo token from the integrated feature \( F_{m} \) to construct the global feature \( F_{G} \in \mathbb{R}^{6 \times C} \).
% Pseudo token only 1 in each modality
Then, the global feature \( F_{G} \) is used as the query, while the sampled local feature \( F_{S} \) serves as the key and value to facilitate interaction:
\begin{equation}
    F_{CDA} = F_{G} + \mathcal{CA}(F_{G}, F_{S}),
\end{equation}
where \( F_{CDA} \in \mathbb{R}^{6 \times C} \) represents the final feature while \( \mathcal{CA} \) denotes the cross attention block.
By incorporating the CDA, we can effectively extract discriminative local information and adaptively aggregate complementary multi-modal features, enhancing the model’s discriminability.
\subsection{Objective Function}
As shown in \textcolor{red}{Fig.}~\ref{fig:Overall}, we optimize IDEA through losses applied to multiple features.
% where \( F_{G} \) is used to optimize IMFE and \( F_{CDA} \) focuses on optimizing CDA.
%
To maintain consistency with prior works~\cite{wang2024top,zhang2024magic}, we separately extract image and text features from \( F_{G} \) and \( F_{CDA} \), resulting in \( F^{v}_{G}, F^{t}_{G} \) and \( F^{v}_{CDA}, F^{t}_{CDA} \in \mathbb{R}^{3C} \).
For each feature, we apply label smoothing cross-entropy loss~\cite{szegedy2016rethinking} and triplet loss~\cite{hermans2017defense}:
\begin{equation}
    \mathcal{L}_{g}(\mathcal{F}) = \mathcal{L}_{CE}(\mathcal{F}) + \mathcal{L}_{Tri}(\mathcal{F}),
\end{equation}
where \( \mathcal{F} \) denotes the input feature.
Here, \( \mathcal{L}_{CE} \) is the label smoothing cross-entropy loss while \( \mathcal{L}_{Tri} \) means the triplet loss.
The overall objective function is then formulated as:
\begin{equation}
    \mathcal{L} = \mathcal{L}_{g}(F^{v}_{G}) + \mathcal{L}_{g}(F^{v}_{CDA}) + \mathcal{L}_{g}(F^{t}_{G}) + \mathcal{L}_{g}(F^{t}_{CDA}).
\end{equation}
\vspace{-8mm} 
\section{Experiments}
\label{sec:experiments}
\begin{table}[t]
  \centering
  \renewcommand\arraystretch{1.12}
  \setlength\tabcolsep{4pt}
  \resizebox{0.42\textwidth}{!}
{
  \begin{tabular}{cccccc}
      \noalign{\hrule height 1pt}
  &\multicolumn{1}{c}{\multirow{2}{*}{\textbf{Methods}}}   & \multicolumn{4}{c}{\textbf{RGBNT201}} \\ \cmidrule(r){3-6}
  & & \textbf{mAP} & \textbf{R-1} & \textbf{R-5} & \textbf{R-10} \\ \hline
  \multirow{14}{*}{\rotatebox{90}{\textbf{Multi-modal}}}
  & HAMNet~\cite{li2020multi}   & 27.7         & 26.3            & 41.5            & 51.7             \\
  & PFNet~\cite{zheng2021robust}    & 38.5         & 38.9            & 52.0            & 58.4             \\
  & IEEE~\cite{wang2022interact}     & 47.5         & 44.4            & 57.1            & 63.6             \\
  & DENet~\cite{zheng2023dynamic}    & 42.4         & 42.2            & 55.3            & 64.5            \\
  & LRMM~\cite{wu2025lrmm} & 52.3 & 53.4 & 64.6 & 73.2\\
  & UniCat$^*$~\cite{crawford2023unicat}   & 57.0         & 55.7            & -            & -            \\
& HTT$^*$~\cite{wang2024heterogeneous} &71.1 &73.4 &83.1 &87.3\\
& TOP-ReID$^*$~\cite{wang2024top}  &72.3 &76.6 &84.7 &89.4\\
& EDITOR$^*$~\cite{zhang2024magic} & 66.5       & 68.3           & 81.1        & 88.2             \\
& RSCNet$^*$~\cite{yu2024representation} & 68.2 & 72.5 & - & - \\
& WTSF-ReID$^*$~\cite{yu2025wtsf} & 67.9 &72.2 &83.4 &89.7 \\
& MambaPro$\dagger$~\cite{wang2024mambapro} & 78.9 & \textbf{83.4} & \underline{89.8} & 91.9 \\
& DeMo$^\dagger$~\cite{wang2024decoupled}  &\underline{79.0} 	 &\underline{82.3} 	 &88.8 	 &\underline{92.0}      \\
\rowcolor[gray]{0.92}
  & $\mathrm{\textbf{IDEA}}^\dagger$  &\textbf{80.2} 	 &82.1 	 &\textbf{90.0} 	 &\textbf{93.3}      \\
  \noalign{\hrule height 1pt}
  \end{tabular}
  }
  \vspace{-1.5mm}
  \caption{Performance comparison on RGBNT201. 
  Best results are in bold, the second bests are underlined. 
  $\dagger$ denotes CLIP-based methods, $*$ indicates ViT-based while others are CNN-based ones.}
  \label{tab:multi-spectral person ReID}
  \vspace{-4mm}
\end{table}
%~~~~~~~~~~~~~~~~~~~~~~~~~~~~~~~~~~~~~~~~~~~~~~~~~~~~~~~~~~~~~~~~~~~~~~~~~~~~~~~~~~~~~~~~~~~~~~~
\subsection{Datasets and Evaluation Protocols}
\textbf{Datasets.}
We evaluate the proposed method on three multi-modal object ReID benchmarks.
To efficiently annotate multi-spectral images without hardware constraints, we employ the API-based Qwen-VL~\cite{bai2023qwen} for automated textual description generation.
To be specific, RGBNT201~\cite{zheng2021robust} comprises 4,787 image triplets with 14,361 annotations, each averaging 35.47 characters and encompassing 8 distinct attributes. 
MSVR310~\cite{zheng2023cross} contains 2,087 image triplets and 6,261 annotations, with an average length of 56.06 characters and 6 attributes. 
RGBNT100~\cite{li2020multi} includes the largest number of triplets at 17,250 and 51,750 annotations with an average length of 56.25 characters and 6 attributes.
\\
\textbf{Evaluation Protocols.}
The performance is evaluated using mean Average Precision (mAP) and Cumulative Matching Characteristics (CMC) at Rank-K (\(K = 1, 5, 10\)). 
\subsection{Implementation Details}
Our model is implemented using PyTorch and trained on an NVIDIA A800 GPU. 
The pre-trained model CLIP~\cite{radford2021learning} is used for both the vision and text encoders. 
For the datasets, images triples are resized to 256$\times$128 for RGBNT201 and 128$\times$256 for MSVR310 and RGBNT100. 
Data augmentation includes random horizontal flipping, cropping and random erasing~\cite{zhong2020random}.
For RGBNT201 and MSVR310, the mini-batch size is set to 64, with 8 images sampled per identity. 
For RGBNT100, the mini-batch size is set to 128, with 16 images sampled per identity.
We fine-tune the learnable modules using an initial learning rate of 3.5$\times$10$^{-6}$, which decays to 3.5$\times$10$^{-7}$ during training.
For the RGBNT201 dataset, the values of \(N_{p}\) and \(k\) are set to 2 and 5, respectively.
Other details are provided in the supplementary material.
\textbf{Code and annotations are available at } \href{https://github.com/924973292/IDEA}{\textbf{IDEA}}.
\begin{table}[t]
    \centering
    \renewcommand\arraystretch{1.2}
    \setlength\tabcolsep{5pt}
    \resizebox{0.42\textwidth}{!}
	{
    \begin{tabular}{cccccc}
        \noalign{\hrule height 1pt}
    &\multicolumn{1}{c}{\multirow{2}{*}{\textbf{Methods}}} &  \multicolumn{2}{c}{\textbf{RGBNT100}} & \multicolumn{2}{c}{\textbf{MSVR310}} \\\cmidrule(r){3-4} \cmidrule(r){5-6}
    & & \textbf{mAP} & \textbf{R-1} & \textbf{mAP} & \textbf{R-1} \\
    \hline
    \multirow{18}{*}{\rotatebox{90}{\textbf{Multi-modal}}}
    &HAMNet~\cite{li2020multi} & 74.5 & 93.3 & 27.1 & 42.3 \\
    &PFNet~\cite{zheng2021robust}& 68.1 & 94.1 & 23.5 & 37.4 \\
    &GAFNet~\cite{guo2022generative} & 74.4 & 93.4 & - & - \\
    &GPFNet~\cite{he2023graph} & 75.0 & 94.5 & - & - \\
    &CCNet~\cite{zheng2023cross} & 77.2 & 96.3 & 36.4 & 55.2 \\
    & LRMM~\cite{wu2025lrmm} & 78.6 & 96.7 & 36.7 &49.7\\
    &GraFT$^*$~\cite{yin2023graft} &76.6 &94.3 &- &-\\
    &UniCat$^*$~\cite{crawford2023unicat}    & 79.4         & 96.2  & -            & -            \\
    &PHT$^*$~\cite{pan2023progressively} & 79.9 & 92.7 & - & - \\
    & HTT$^*$~\cite{wang2024heterogeneous} &75.7&92.6&- &-\\
    & TOP-ReID$^*$~\cite{wang2024top} &81.2 & 96.4 & 35.9 & 44.6 \\
    & EDITOR$^*$~\cite{zhang2024magic} & 82.1 & 96.4 &39.0 & 49.3\\
    & FACENet$^*$~\cite{zheng2025flare} & 81.5 &\underline{96.9} &36.2 &54.1 \\
    & RSCNet$^*$~\cite{yu2024representation} &82.3 &96.6 &39.5 &49.6\\
    & WTSF-ReID$^*$~\cite{yu2025wtsf} & 82.2 &96.5 & 39.2 & 49.1 \\
    & MambaPro$\dagger$~\cite{wang2024mambapro} & 83.9 & 94.7 & \underline{47.0} & 56.5 \\
    & DeMo$\dagger$~\cite{wang2024decoupled} &\underline{86.2} 	&\textbf{97.6} &\textbf{49.2}	&\underline{59.8} \\
    \rowcolor[gray]{0.92}
    & $\mathrm{\textbf{IDEA}}^\dagger$& \textbf{87.2} 	&96.5 &\underline{47.0}	&\textbf{62.4} \\
    \noalign{\hrule height 1pt}
    \end{tabular}
    }
    \vspace{-1.5mm}
    \caption{Performance on RGBNT100 and MSVR310.}
    \label{tab:multi-spectral vehicle ReID}
    \vspace{-4mm}
\end{table}
%~~~~~~~~~~~~~~~~~~~~~~~~~~~~~~~~~~~~~~~~~~~~~~~~~~~~~~~~~~~~~~~~~~~~~~~~~~~~~~~~~~~~~~~~~~~~~~~
\subsection{Comparison with State-of-the-Art Methods}
\textbf{Multi-modal Person ReID.}
In \textcolor{red}{Tab.}~\ref{tab:multi-spectral person ReID}, we compare our method IDEA$^\dagger$ with existing multi-modal approaches on the RGBNT201 dataset. 
Leveraging complementary information from different modalities, multi-modal methods exhibit superior performance.
Specifically, our proposed IDEA$^\dagger$ achieves 80.2\% mAP and 82.1\% Rank-1 accuracy, surpassing TOP-ReID$^*$ by 7.9\% in mAP and 5.5\% in Rank-1. 
Compared with other leading methods like HTT$^*$ and EDITOR$^*$, IDEA$^\dagger$ shows a superior adaptability to complex scenarios, confirming the robustness.
Besides, IDEA$^\dagger$ outperforms the CLIP-based methods MambaPro$^\dagger$ and DeMo$^\dagger$ by 1.3\% and 1.2\% in mAP, respectively.
These results highlight IDEA’s ability to utilize semantic information from textual guidance for improved feature discrimination. 
\\
\textbf{Multi-modal Vehicle ReID.}
In \textcolor{red}{Tab.}~\ref{tab:multi-spectral vehicle ReID}, we compare IDEA$^\dagger$ with state-of-the-art methods on the RGBNT100 and MSVR310 datasets.
Our IDEA$^\dagger$ achieves an mAP of 87.2\%, surpassing EDITOR$^*$ by 5.1\% in mAP.
Especially on the challenging MSVR310 dataset, IDEA$^\dagger$ achieves an mAP of 47.0\% and a Rank-1 accuracy of 62.4\%, outperforming EDITOR$^*$ by 8.0\% in mAP and 13.1\% in Rank-1 accuracy.
These results verify IDEA's generalization ability.

%~~~~~~~~~~~~~~~~~~~~~~~~~~~~~~~~~~~~~~~~~~~~~~~~~~~~~~~~~~~~~~~~~~~~~~~~~~~~~~~~~~~~~~~~~~~~~~~
\begin{table}[t]
  \centering
  \renewcommand\arraystretch{1.0}
  \setlength\tabcolsep{4.5pt}
  \resizebox{0.35\textwidth}{!}
  {
  \begin{tabular}{cccccc}
      \noalign{\hrule height 1pt}
      \multicolumn{1}{c}{\multirow{2}{*}{\textbf{Index}}} &\multicolumn{3}{c}{\textbf{Modules}} & \multicolumn{2}{c}{\textbf{Metrics}} \\
      \cmidrule(r){2-4} \cmidrule(r){5-6}
 & \textbf{Text}              & \textbf{IMFE}                & \textbf{CDA}                   & \textbf{mAP}    & \textbf{Rank-1}   \\\hline
  A                  & \ding{53}                  & \ding{53}                  & \ding{53}                    & 70.3  & 72.1 \\
  B                  & \ding{51}                  & \ding{53}                  & \ding{53}                      & 73.4  & 75.8 \\
  \multirow{1}{*}{C} & \multirow{1}{*}{\ding{51}} & \multirow{1}{*}{\ding{51}} & \multirow{1}{*}{\ding{53}}    & 77.2  & 81.1 \\
  \rowcolor[gray]{0.92}
  \multirow{1}{*}{D} & \multirow{1}{*}{\ding{51}} & \multirow{1}{*}{\ding{51}} & \multirow{1}{*}{\ding{51}}    &\textbf{80.2} &\textbf{82.1}  \\
  \noalign{\hrule height 1pt}
  \end{tabular}
  }
  \vspace{-1.5mm}
  \caption{Performance comparison with different modules.}
  \label{tab:main_ablation}
  \vspace{-2mm}
\end{table}
%~~~~~~~~~~~~~~~~~~~~~~~~~~~~~~~~~~~~~~~~~~~~~~~~~~~~~~~~~~~~~~~~~~~~~~~~~~~~~~~~~~~~~~~~~~~~~~~
\begin{table}[t]
  \centering
  \renewcommand\arraystretch{1.0}
  \setlength\tabcolsep{4.5pt}
  \resizebox{0.42\textwidth}{!}
  {
  \begin{tabular}{cccccc}
      \noalign{\hrule height 1pt}
      \multicolumn{1}{c}{\multirow{2}{*}{\textbf{Index}}} &\multicolumn{3}{c}{\textbf{IMFE}} & \multicolumn{2}{c}{\textbf{Metrics}} \\
      \cmidrule(r){2-4} \cmidrule(r){5-6}
 & \textbf{InverseNet}              & \textbf{Prefixes}                & \textbf{Prompt}                   & \textbf{mAP}    & \textbf{Rank-1}   \\\hline
  A                   & \ding{53}                  & \ding{53}                 & -                     & 72.6 & 75.1  \\
  B                  & \ding{53}                  & \ding{51}                  & \ding{53}                     & 73.4  & 75.8                   \\
  C                  & \ding{51}                  & \ding{53}                  & -                    & 73.7  & 77.3                   \\
  D                  & \ding{51}                  & \ding{51}                  & \ding{53}                      & 75.4  & 78.6               \\
  \rowcolor[gray]{0.92}
  E & \multirow{1}{*}{\ding{51}} & \multirow{1}{*}{\ding{51}} & \multirow{1}{*}{\ding{51}}    & \textbf{77.2}  & \textbf{81.1}  \\
  \noalign{\hrule height 1pt}
  \end{tabular}
  }
  \vspace{-1.5mm}
  \caption{Comparison with different components in IMFE.}
  \label{tab:IMFE_ablation}
  \vspace{-4mm}
\end{table}
%~~~~~~~~~~~~~~~~~~~~~~~~~~~~~~~~~~~~~~~~~~~~~~~~~~~~~~~~~~~~~~~~~~~~~~~~~~~~~~~~~~~~~~~~~~~~~~~
\subsection{Ablation Studies}
\begin{figure}[t]
  \centering
    \resizebox{0.475\textwidth}{!}
    {
  \includegraphics[width=1.\linewidth]{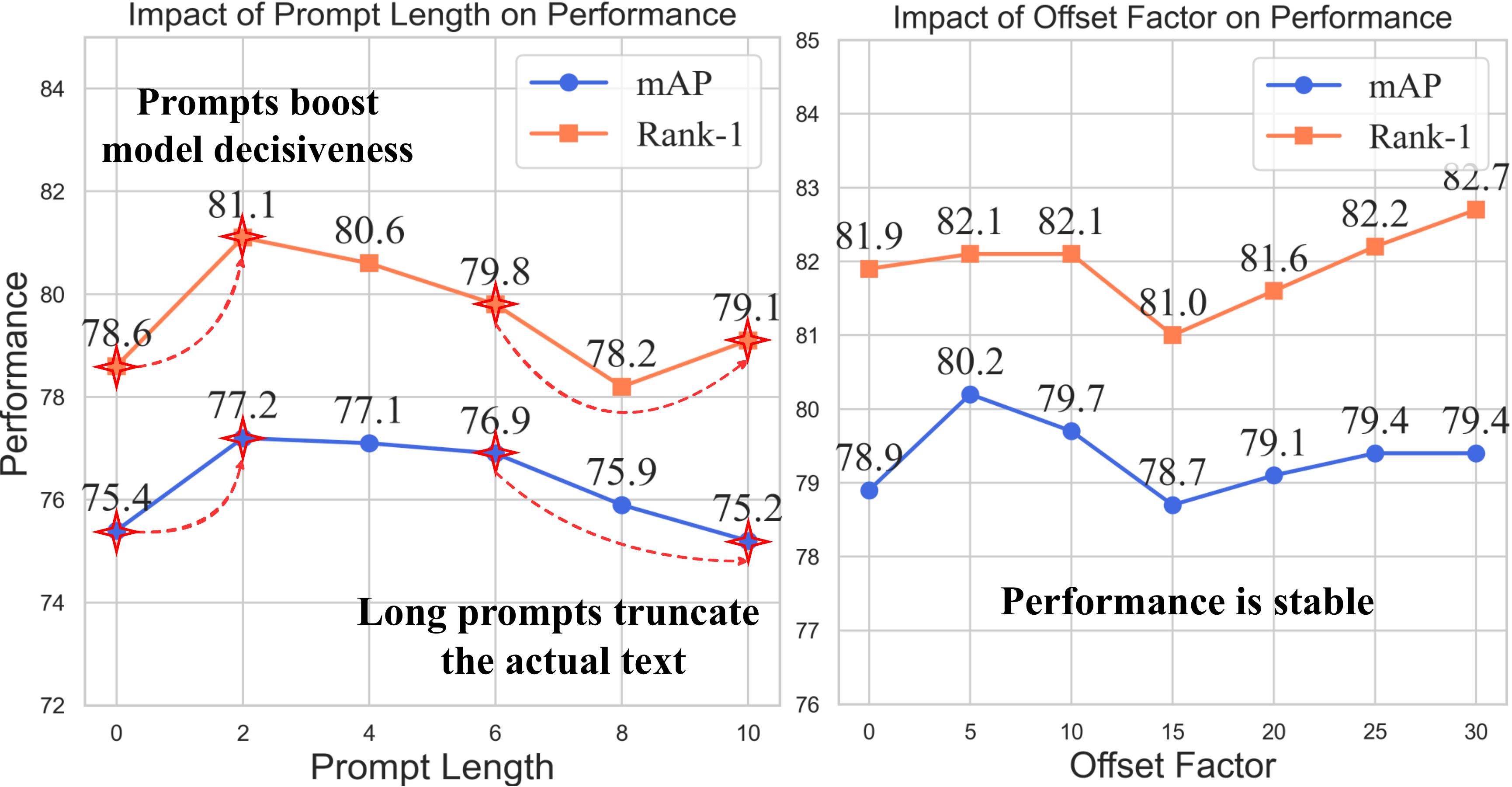}
  }
  \vspace{-5mm}
   \caption{Comparison with different hyper-parameters.}
  \label{fig:hyper}
  \vspace{-2mm}
\end{figure}

We evaluate the effectiveness of the proposed modules on the RGBNT201 dataset. 
To be specific, our baseline comprises a three-branch vision encoder, which performs retrieval by concatenating the class tokens from three modalities. 
Upon incorporating the IMFE, we utilize \( F^{t}_{G} \) for retrieval and when the CDA is included, \( F^{t}_{CDA} \) is employed.
\\
\textbf{Effects of Key Modules.}
\textcolor{red}{Tab.}~\ref{tab:main_ablation} presents the performance of various combinations of our proposed modules. 
Model A serves as the baseline, achieving an mAP of 70.3\% and Rank-1 accuracy of 72.1\%.
Model B incorporates text information through parallel text encoders, improving the mAP to 73.4\% and Rank-1 accuracy to 75.8\%. 
With the inverted structure in IMFE, Model C further enhances performance, reaching an mAP of 77.2\% and Rank-1 of 81.1\%. 
Finally, Model D integrates CDA, delivering the best performance with an mAP of 80.2\%.
These results fully validate the effectiveness and efficiency of the proposed modules.
%~~~~~~~~~~~~~~~~~~~~~~~~~~~~~~~~~~~~~~~~~~~~~~~~~~~~~~~~~~~~~~~~~~~~~~~~~~~~~~~~~~~~~~~~~~~~~~~
\begin{table}[t]
  \centering
  \renewcommand\arraystretch{1.0}
  \setlength\tabcolsep{4.5pt}
  \resizebox{0.42\textwidth}{!}
  {
  \begin{tabular}{cccccc}
      \noalign{\hrule height 1pt}
      \multicolumn{1}{c}{\multirow{2}{*}{\textbf{Index}}} &\multicolumn{3}{c}{\textbf{CDA}} & \multicolumn{2}{c}{\textbf{Metrics}} \\
      \cmidrule(r){2-4} \cmidrule(r){5-6}
& \textbf{Sample}              & \textbf{Cross Attn}                & \textbf{Shared Offset}                   & \textbf{mAP}    & \textbf{Rank-1}   \\\hline
A                   & \ding{53}                  & \ding{53}                  & - & 76.3  & 78.7   \\
B                  & \ding{53}                  & \ding{51}                  & -                     & 77.0  & 79.8                    \\
C                  & \ding{51}                  & \ding{53}                  & \ding{53}                    & 76.8  & 78.9                    \\
D                  & \ding{51}                  & \ding{53}                  & \ding{51}                      & 77.6  & 80.4                \\
E                  & \ding{51}                  & \ding{51}                  & \ding{53}                      & 79.5  & 81.7                \\
\rowcolor[gray]{0.92}
F & \multirow{1}{*}{\ding{51}} & \multirow{1}{*}{\ding{51}} & \multirow{1}{*}{\ding{51}}     &\textbf{80.2}  &\textbf{82.1}   \\
  \noalign{\hrule height 1pt}
  \end{tabular}
  }
  \vspace{-1.5mm}
  \caption{Comparison with different components in CDA.}
  \label{tab:CDA_ablation}
  \vspace{-3mm}
\end{table}
%~~~~~~~~~~~~~~~~~~~~~~~~~~~~~~~~~~~~~~~~~~~~~~~~~~~~~~~~~~~~~~~~~~~~~~~~~~~~~~~~~~~~~~~~~~~~~~~
\begin{table}[t]
  \centering
  \renewcommand\arraystretch{1.0}
  \setlength\tabcolsep{4.5pt}
  \resizebox{0.42\textwidth}{!}
  {
  \begin{tabular}{cccccc}
      \noalign{\hrule height 1pt}
      \multicolumn{1}{c}{\textbf{Model}} & \textbf{mAP} & \textbf{Rank-1} & \textbf{Rank-5} & \textbf{Rank-10} \\ \hline
  IDEA w/o Text  & 74.5 & 75.0 & 84.8 & 88.8 \\
  \rowcolor[gray]{0.92}
  IDEA            & \textbf{80.2} & \textbf{82.1} & \textbf{90.0} & \textbf{93.3} \\
      \noalign{\hrule height 1pt}
  \end{tabular}
  }
  \vspace{-1.5mm}
  \caption{Comparison of IDEA with and without text.}
  \label{tab:Text_ablation}
  \vspace{-3mm}
\end{table}
%~~~~~~~~~~~~~~~~~~~~~~~~~~~~~~~~~~~~~~~~~~~~~~~~~~~~~~~~~~~~~~~~~~~~~~~~~~~~~~~~~~~~~~~~~~~~~~~
\begin{table}[t]
  \centering
  \renewcommand\arraystretch{1.0}
  \setlength\tabcolsep{4.5pt}
  \resizebox{0.42\textwidth}{!}
  {
  \begin{tabular}{cccccccc}
      \noalign{\hrule height 1pt}
      \multicolumn{1}{c}{\textbf{Model}} & \textbf{mAP} & \textbf{Rank-1} & \textbf{Rank-5} & \textbf{Rank-10} \\ \hline
  IDEA w/o Offset  & 78.4 & 81.3 & 89.7 & 92.2 \\
  \rowcolor[gray]{0.92}
  IDEA            & \textbf{80.2} & \textbf{82.1} & \textbf{90.0} & \textbf{93.3} \\
      \noalign{\hrule height 1pt}
  \end{tabular}
  }
  \vspace{-1.5mm}
  \caption{Comparison of IDEA with and without offset.}
  \label{tab:Offset_ablation}
  \vspace{-3mm}
\end{table}
%~~~~~~~~~~~~~~~~~~~~~~~~~~~~~~~~~~~~~~~~~~~~~~~~~~~~~~~~~~~~~~~~~~~~~~~~~~~~~~~~~~~~~~~~~~~~~~~
\\
\textbf{Effects of Key Components in IMFE.}
\textcolor{red}{Tab.}~\ref{tab:IMFE_ablation} shows the performance of different components in the IMFE.
Model A represents the parallel text encoders without the inverted structure or prefixes, achieving an mAP of 72.6\% and Rank-1 accuracy of 75.1\%.
With the Modal Prefixes in Model B, the mAP increases to 73.4\%, demonstrating the effectiveness of our prefixes mechanism.
Model C introduces the inverted structure, achieving an mAP of 73.7\%.
Model D combines the inverted structure and prefixes, further improving the mAP to 75.4\%.
Finally, Model E integrates the learnable tokens in Modal Prefixes, achieving the best performance with an mAP of 77.2\%.
Overall, inverted structure performs better than the parallel structure, achieving better performance.
These results demonstrate the effectiveness of different components in our proposed IMFE.
\\
\textbf{Effects of Key Components in CDA.}
Based on the IMFE, \textcolor{red}{Tab.}~\ref{tab:CDA_ablation} shows the performance of various components within CDA.
When sampling is excluded, all local patches in \( \hat{F}_{m} \) are processed.
Without cross attention, multi-modal interaction uses only local features, omitting global context, with pooled \( \hat{F}_{m} \) for retrieval.
If shared offset is excluded, each modality independently generates offsets.
Regarding numerical results, Model A, which does not incorporate these components, achieves an mAP of 76.3\% and Rank-1 accuracy of 78.7\%.
Model B introduces cross attention, leading to a slight improvement in mAP to 77.0\%.
Model C adds sampling, resulting in a 0.5\% improvement in mAP over Model A by reducing noise in the local features.
Model D incorporates shared offset, yielding a 0.8\% improvement in mAP and a 1.5\% increase in Rank-1 accuracy compared to Model C.
Model E combines cross attention and sampling, achieving an mAP of 79.5\%.
Finally, Model F delivers the best performance with an mAP of 80.2\%.
These results validate the effectiveness of each component within CDA.
\\
\textbf{Effects of Text Guidance in IDEA.}
\textcolor{red}{Tab.}~\ref{tab:Text_ablation} compares IDEA's performance with and without text input.
Model A, lacking text input, achieves an mAP of 74.5\%. 
With text guidance, Model B improves to 80.2\% mAP, demonstrating the effectiveness of text in enhancing feature robustness.
\\
\textbf{Effects of Offset in IDEA.}
\textcolor{red}{Tab.}~\ref{tab:Offset_ablation} compares the performance of IDEA with and without the offset mechanism.
Model A uses aggregated local features with convolution to interact with global features, achieving an mAP of 78.4\% and Rank-1 accuracy of 81.3\%.
Model B introduces the offset mechanism, improving the mAP to 80.2\% and Rank-1 accuracy to 82.1\%. 
This demonstrates that the offset mechanism helps the model focus on more important regions.
\\
\textbf{Effects of Prompt Lengths and Offset Factors.}
\textcolor{red}{Fig.}~\ref{fig:hyper} illustrates the performance of IDEA with varying prompt lengths and offset factors. 
The results show that a prompt length of 2 and an offset factor of 5 yield the best performance.
As the prompt length increases, the model's performance improves, but the gains become smaller as the prompt length grows, likely due to the text truncation.
Regarding the offset factor, the model remains stable.
%~~~~~~~~~~~~~~~~~~~~~~~~~~~~~~~~~~~~~~~~~~~~~~~~~~~~~~~~~~~~~~~~~~~~~~~~~~~~~~~~~~~~~~~~~~~~~~~
%~~~~~~~~~~~~~~~~~~~~~~~~~~~~~~~~~~~~~~~~~~~~~~~~~~~~~~~~~~~~~~~~~~~~~~~~~~~~~~~~~~~~~~~~~~~~~~~
\begin{figure}[t]
  \centering
    \resizebox{0.475\textwidth}{!}
    {
  \includegraphics[width=30.\linewidth]{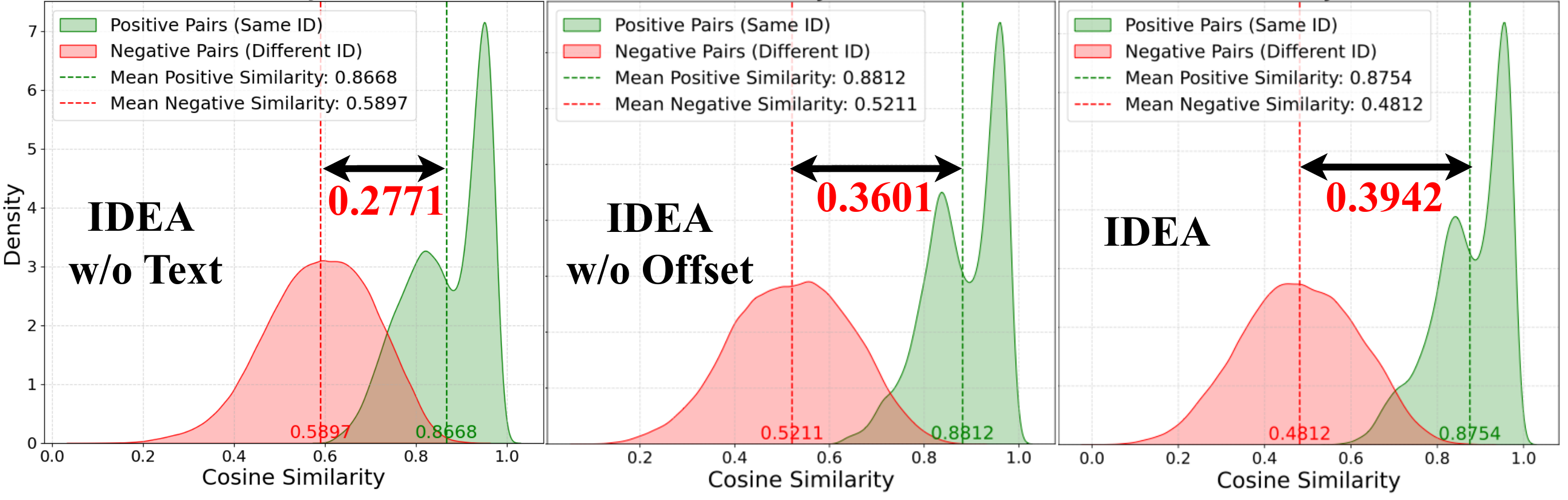}
  }
  \vspace{-5mm}
   \caption{Visualization of the cosine similarity distribution.}
  \label{fig:cosine}
  \vspace{-4mm}
\end{figure}
%~~~~~~~~~~~~~~~~~~~~~~~~~~~~~~~~~~~~~~~~~~~~~~~~~~~~~~~~~~~~~~~~~~~~~~~~~~~~~~~~~~~~~~~~~~~~~~~
\begin{figure*}[t]
  \centering
    \resizebox{0.96\textwidth}{!}
    {
  \includegraphics[width=1\linewidth]{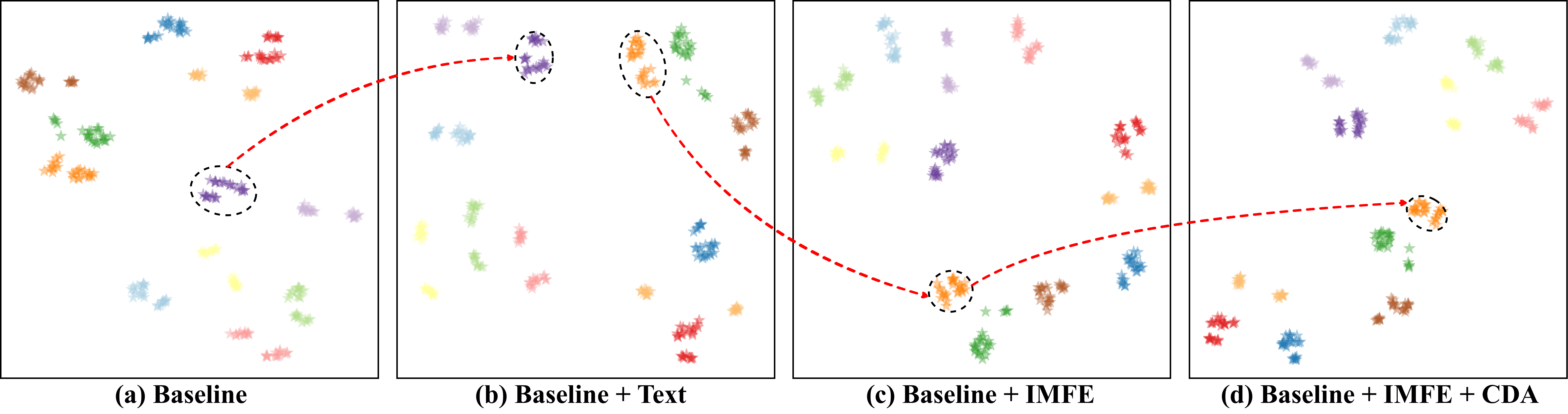}
  }
  \vspace{-2mm}
   \caption{Visualization of the feature distributions with t-SNE~\cite{van2008visualizing}.
   Different colors represent different identities.}
  \label{fig:tsne}
  \vspace{-5mm}
\end{figure*}
%~~~~~~~~~~~~~~~~~~~~~~~~~~~~~~~~~~~~~~~~~~~~~~~~~~~~~~~~~~~~~~~~~~~~~~~~~~~~~~~~~~~~~~~~~~~~~~~
\begin{figure}[t]
  \centering
    \resizebox{0.45\textwidth}{!}
    {
  \includegraphics[width=1.\linewidth]{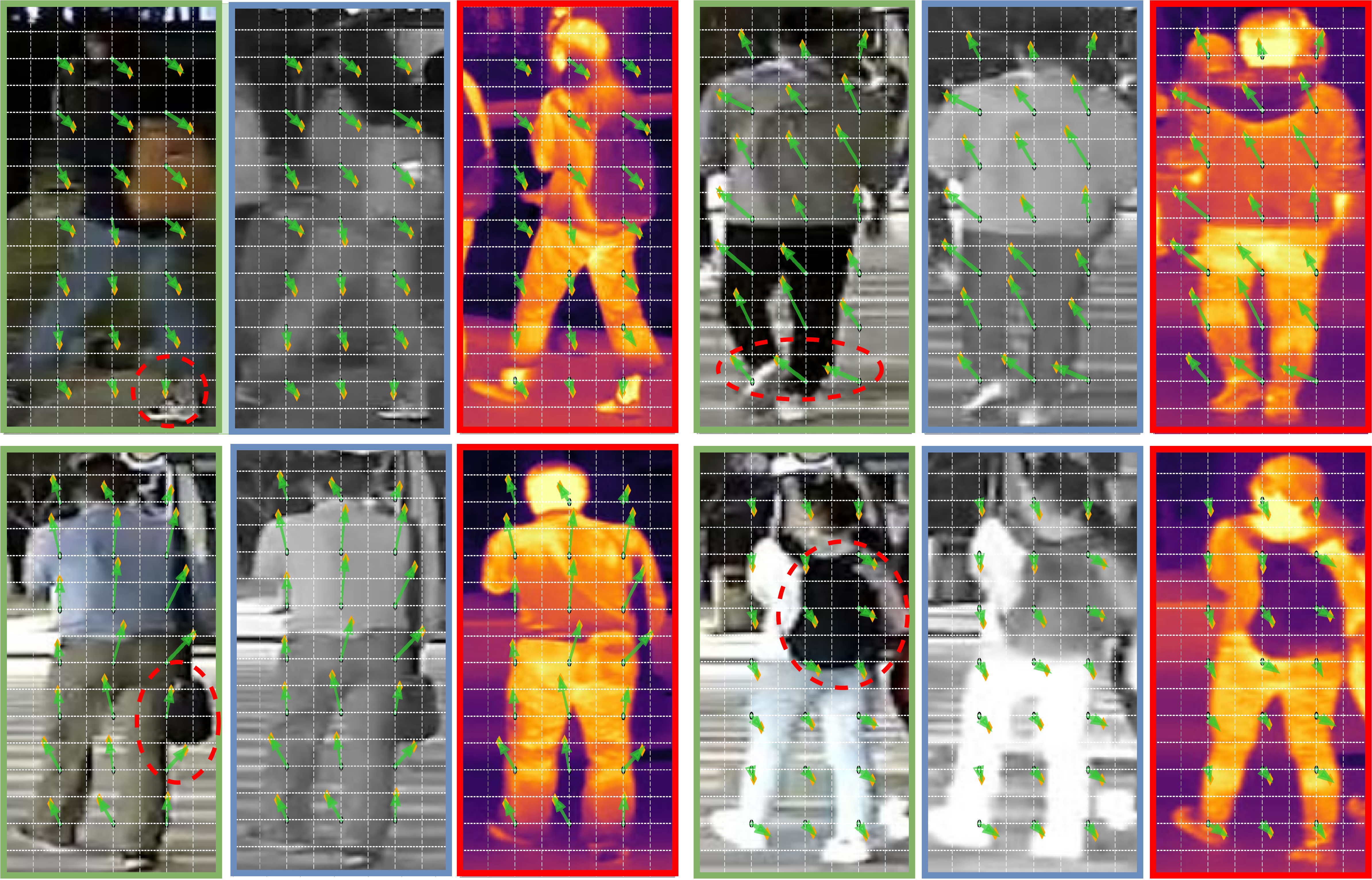}
  }
  \vspace{-2mm}
   \caption{Visualization of the generated offsets.}
  \label{fig:off_set}
  \vspace{-4mm}
\end{figure}
%~~~~~~~~~~~~~~~~~~~~~~~~~~~~~~~~~~~~~~~~~~~~~~~~~~~~~~~~~~~~~~~~~~~~~~~~~~~~~~~~~~~~~~~~~~~~~~~
\begin{figure}[t]
  \centering
    \resizebox{0.45\textwidth}{!}
    {
  \includegraphics[width=30\linewidth]{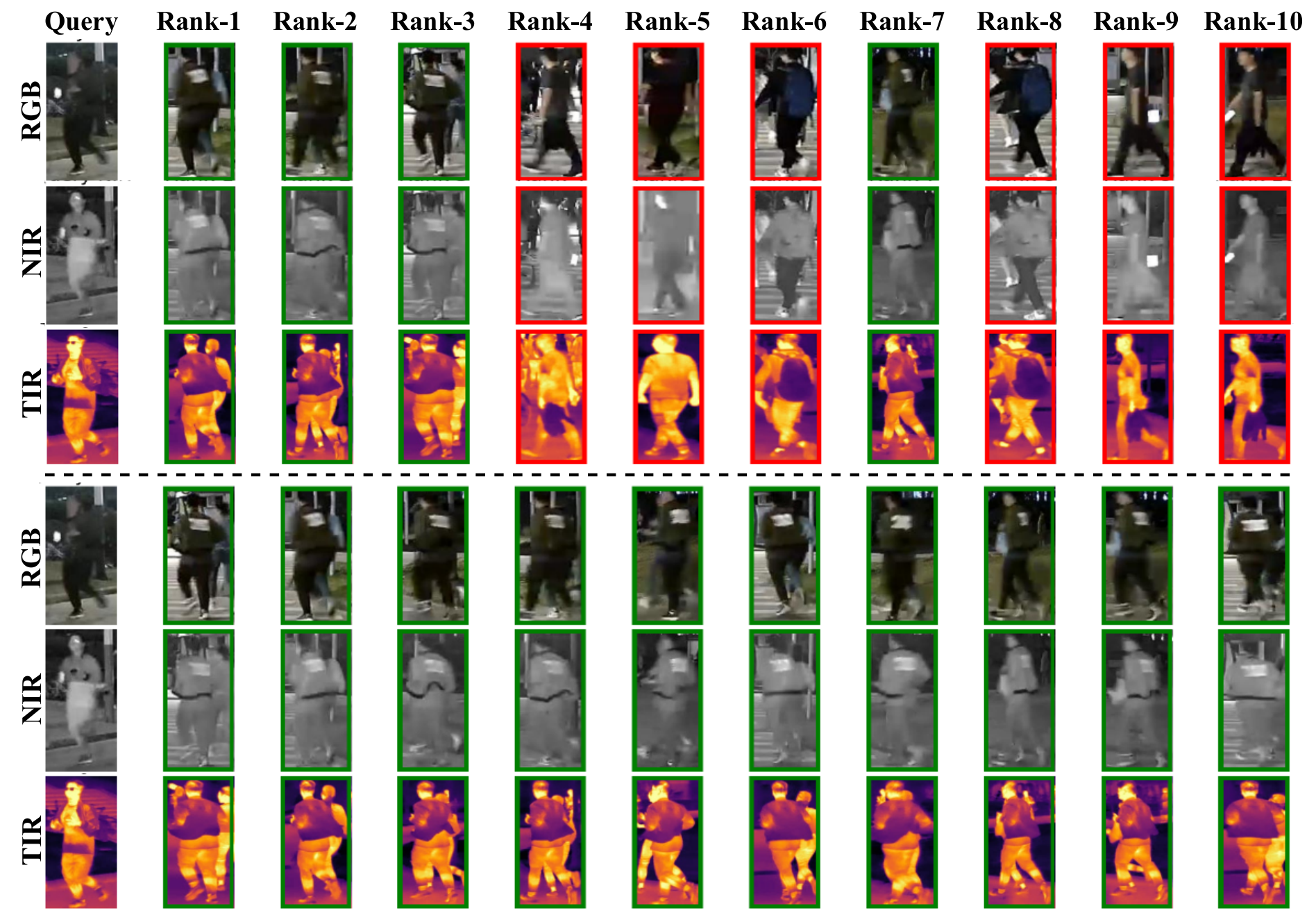}
  }
  \vspace{-2mm}
   \caption{Rank list comparison between the baseline and IDEA.}
  \label{fig:rank_list}
  \vspace{-5mm}
\end{figure}
%~~~~~~~~~~~~~~~~~~~~~~~~~~~~~~~~~~~~~~~~~~~~~~~~~~~~~~~~~~~~~~~~~~~~~~~~~~~~~~~~~~~~~~~~~~~~~~~
\begin{figure}[t]
  \centering
    \resizebox{0.45\textwidth}{!}
    {
  \includegraphics[width=30\linewidth]{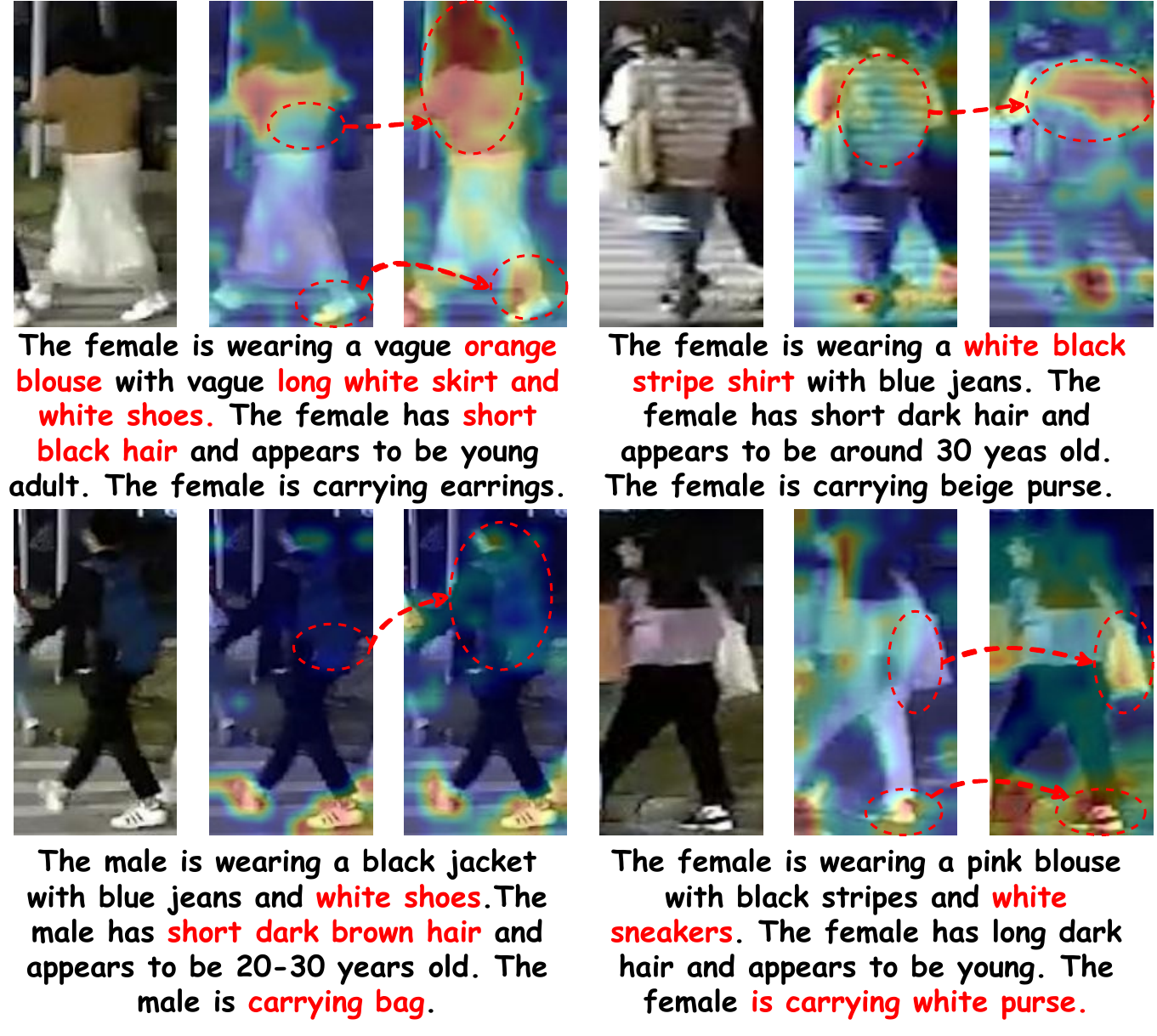}
  }
  \vspace{-1mm}
   \caption{Visualization of the channel activation maps.
   Each set includes the original image, baseline and IDEA map, respectively.}
  \label{fig:text_gain}
  \vspace{-5mm}
\end{figure}
\subsection{Visualization Analysis}
\textbf{Cosine Similarity Distributions.}
\textcolor{red}{Fig.}~\ref{fig:cosine} presents the distributions of cosine similarities among test features.
The results verify that incorporating text enhances feature discrimination, while adding offset mechanism further amplifies the distinction between positive and negative samples, confirming the capture of more discriminative features.
\\
\textbf{Multi-modal Feature Distributions.}
\textcolor{red}{Fig.}~\ref{fig:tsne} visualizes the feature distributions of different modules.
Comparing \textcolor{red}{Fig.}~\ref{fig:tsne} (a) and (b), the introduction with text guidance improves feature discrimination, as instances of the same ID become more compact.
In \textcolor{red}{Fig.}~\ref{fig:tsne} (c), with IMFE, the feature distributions become more discriminative than the parallel structure in \textcolor{red}{Fig.}~\ref{fig:tsne} (b).
Finally, in \textcolor{red}{Fig.}~\ref{fig:tsne} (d), CDA further enhances feature discrimination.
These visualizations demonstrate the effectiveness of our proposed modules.
\\
\textbf{Visualization of the Generated Offsets.}
\textcolor{red}{Fig.}~\ref{fig:off_set} visualizes the offsets generated by CDA.
Each arrow starts at the reference point and ends at the feature sampling point.
We map these offsets onto the original image to identify the regions where the model shifts its attention.
As shown, the offsets highlight discriminative areas, such as the head, bag and shoes. 
It demonstrates how the offset mechanism guides the model to focus on the most discriminative regions, effectively reducing the influence of noisy information.
\\
\textbf{Rank List Comparison.}
\textcolor{red}{Fig.}~\ref{fig:rank_list} compares the cross-camera rank lists from the baseline and IDEA.
IDEA produces more accurate rankings, while the baseline yields noisier results.
These visualizations validate the effectiveness of IDEA.
\\
\textbf{Visualization of Channel Activation Maps.}
\textcolor{red}{Fig.}~\ref{fig:text_gain} compares the channel activation maps of our baseline model and IDEA.
The incorporation of textual guidance helps the model focus on more discriminative regions, enhancing feature robustness and improving interpretability.

\section{Conclusion}
\label{sec:conclusion}
In this work, we propose IDEA, a novel feature learning framework for multi-modal object ReID.
We first construct three text-enhanced multi-modal object ReID benchmarks using MLLMs, providing a structured caption generation pipeline.
With the generated text, the Inverted Multi-modal Feature Extractor (IMFE) leverages semantic guidance from inverted texts while reducing fusion conflicts.
Additionally, the Cooperative Deformable Aggregation (CDA) adaptively aggregates discriminative local features with global information.
Experiments on three public ReID benchmarks demonstrate the effectiveness of our method.
\\
{\small
\textbf{Acknowledgements.}
This work was supported in part by the National Natural Science Foundation of China (No. 62101092, 62476044, 62388101) and Open Project of Anhui Provincial Key Laboratory of Multimodal Cognitive Computation, Anhui University (No. MMC202102, MMC202407).} % 这里结束 \small 的作用范围
\vspace*{-4mm} 
{
    \small
    \bibliographystyle{ieeenat_fullname}
    \bibliography{cvpr25}
}
\clearpage
\setcounter{page}{1}
\maketitlesupplementary
\section{Introduction}
In this supplementary material, we provide comprehensive experimental details, visual examples and extended analyses to support the findings of the main manuscript.
To be specific, the supplementary material is organized as follows:
\begin{enumerate}
\item \textbf{Detailed description of our caption generation pipeline and more examples from different datasets:}
\begin{itemize}
    \item Detailed description of the caption generation pipeline
    \item More examples from the person/vehicle ReID datasets
\end{itemize}
\item \textbf{Details of the proposed modules and experiments:}
\begin{itemize}
    \item Implementation details of the proposed modules
    \item More explanations of models in the ablation study
\end{itemize}
\item \textbf{Module validation and hyper-parameter analysis:}
\begin{itemize}
    \item Analysis of model parameters and performance
    \item Comparison with different inverse directions in IMFE
    \item Comparison with CLIP-based method
    \item Training efficiency comparison
    \item Effect of text with different modalities combination
    \item Exploration of extreme cases with missing text
    \item Effect of key modules on vehicle dataset
    \item Hyper-parameter analysis for the vehicle dataset
\end{itemize}
\item \textbf{Visualization analysis of IDEA:}
\begin{itemize}
    \item Visualization of multi-modal ranking list
    \item Visualization of the channel activation maps
\end{itemize}
\end{enumerate}
These analyses provide a deeper understanding of our constructed datasets, the proposed modules and experimental results, further validating the effectiveness of our method.
\section{Multi-modal Caption Generation}
\subsection{Details of the Caption Generation Pipeline}
To bridge the gap in multi-modal object ReID captioning, we propose a novel caption generation pipeline that leverages MLLMs to generate informative text descriptions.
Specifically, our pipeline consists of two steps: (1) Caption Generation and (2) Attribute Extraction.
Without loss of generality, we take the multi-modal person ReID dataset as an example to illustrate the caption generation pipeline.
\textbf{Caption Generation:}
The unique characteristics of multi-spectral images and the limitations of MLLMs in captioning make simultaneous annotation of paired multi-spectral images challenging.
When presented with images from multiple spectra, MLLMs often focus on shared semantic information while neglecting modality-specific details.
To address this, we design modality-specific templates to annotate each spectrum independently, ensuring rich and detailed semantic descriptions.
Specifically, this caption generation step offers two main advantages.
First, it enhances modality-specific descriptions by simplifying the captioning task, enabling the model to focus on detailed attributes unique to each spectrum.
Second, it broadens applicability by supporting more cases such as text-to-image ReID in the infrared spectrum, which is particularly advantageous in challenging visual environments.
Technically, our modality-specific templates consist of three key components: a modality-specific prefix, a generic annotation template and an anti-hallucination instruction.
\begin{enumerate}
    \item \textbf{Modality-specific Prefix:} Directs the model to focus on a particular spectrum (e.g., RGB, NIR, or TIR), providing a detailed prompt to describe key attributes such as clothing, hairstyle and belongings:
    \begin{quote}
      \small
      \textit{“Write a comprehensive description of the person's overall appearance based on the [RGB/NIR/TIR] image, strictly following this template. Include the following attributes: 'upper garment', 'lower garment', 'shoes', 'hairstyle', 'gender', 'age group' and 'belongings'. Use specific details, including color, patterns and texture details. Please follow this structure: ”}
    \end{quote}
    \item \textbf{Generic Annotation Template:} Ensures consistency across captions with the following structure:
    \begin{quote}
      \small
    \textit{“The \{Gender\} is wearing a \{Upper\_Color\} \{Upper\_Garment\} with \{Lower\_Color\} \{Lower\_Garment\} and \{Shoe\_Color\} \{Shoes\}. The \{Gender\} has \{Hairstyle\} \{Hair\_Color\} hair and appears to be \{Age\_Group\}. The \{Gender\} is carrying \{Belongings\}."}
    \end{quote}
    \item \textbf{Anti-hallucination Instruction:} Prevents imagined details~\cite{tan2024harnessing} by explicitly guiding the model to focus only on visible attributes:
    \begin{quote}
      \small
    \textit{“If certain attributes are not visible, ignore them. Do not imagine contents not present in the image. Adhere strictly to the format without adding extra explanations."}
    \end{quote}
\end{enumerate}
By combining images from specific spectra with their corresponding templates, MLLMs generate tailored and comprehensive text descriptions for each modality.
This approach provides detailed semantic information and effectively addresses the challenges of multi-modal data annotation.
\begin{figure*}[t]
  \centering
    \resizebox{0.92\textwidth}{!}
    {
  \includegraphics[width=1.\linewidth]{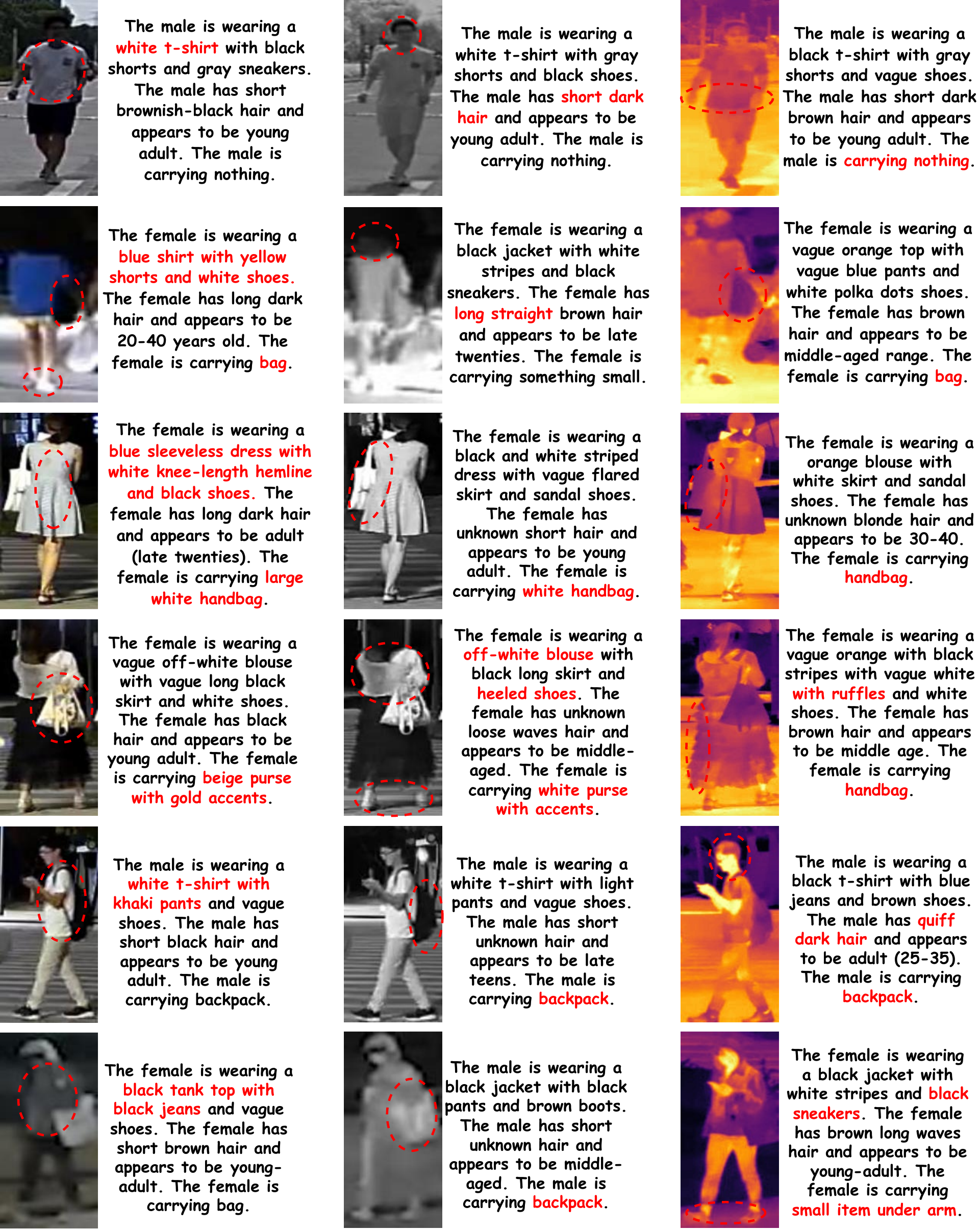}
  }
  \vspace{-2mm}
   \caption{More examples from the multi-modal person ReID dataset RGBNT201.}
  \label{fig:more_instance_person}
  \vspace{-6mm}
\end{figure*}
%～～～～～～～～～～～～～～～～～～～～～～～～～～～～～～～～
\begin{figure*}[t]
  \centering
    \resizebox{0.96\textwidth}{!}
    {
  \includegraphics[width=1.\linewidth]{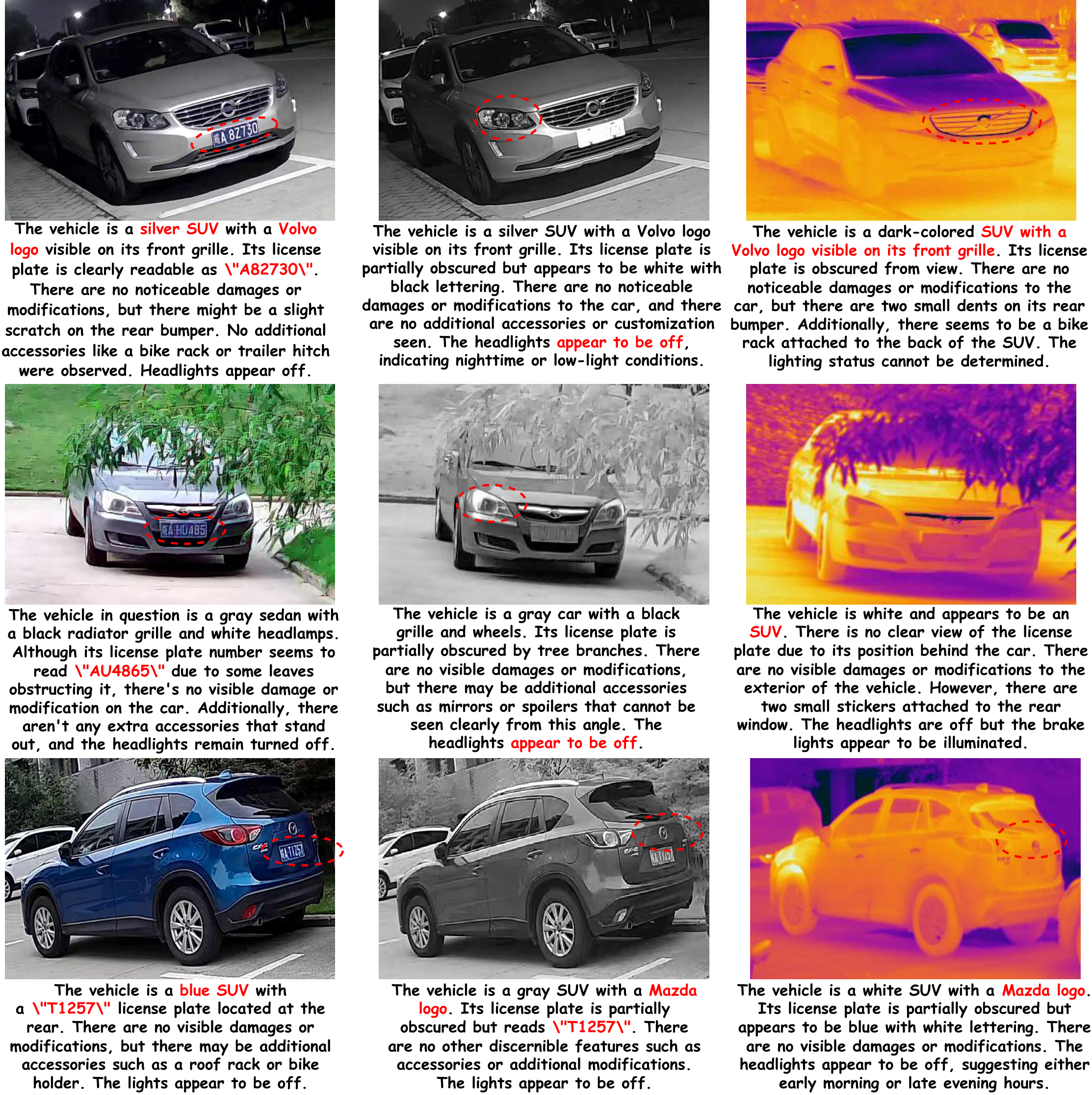}
  }
  \vspace{-2mm}
   \caption{More examples from the multi-modal vehicle ReID dataset MSVR310.}
  \label{fig:more_instance_vehicle}
  \vspace{-6mm}
\end{figure*}
%～～～～～～～～～～～～～～～～～～～～～～～～～～～～～～～～
\\
\textbf{Attribute Extraction:}
During the Caption Generation phase, we generate informative text descriptions for each spectrum. However, due to the inherent randomness of MLLMs during the annotation process, issues such as disorganized structure and overly verbose descriptions persist, even with the use of strict templates during generation.
To address these challenges, we initially use regular expressions to extract key information from the sentences. While this approach captures most essential attributes, it struggles with handling diverse and complex sentence structures. Fortunately, MLLMs exhibit robust capabilities in extracting key information from text.
Leveraging this, we feed the generated descriptions back into the MLLMs with an attribute extraction prompt to specify the predefined attributes to be extracted.
The extracted information is then mapped into the template we use in the caption generation phase:
\begin{quote}
  \small
\textit{“Extract the key attributes from the sentence I give you and fill them into the following template: The \{Gender\} is wearing a \{Upper\_Color\} \{Upper\_Garment\} with \{Lower\_Color\} \{Lower\_Garment\} and \{Shoe\_Color\} \{Shoes\}. The \{Gender\} has \{Hairstyle\} \{Hair\_Color\} hair and appears to be \{Age\_Group\}. The \{Gender\} is carrying \{Belongings\}. Strictly follow the template, do not add any extra information."}
\end{quote}
This process ensures the generation of concise and informative textual descriptions, effectively mitigating the inconsistencies introduced during the initial phase.
\subsection{More Examples from Different Datasets}
\textbf{Multi-modal Person ReID Dataset Examples.}
In \textcolor{red}{Fig.}~\ref{fig:more_instance_person}, we present additional examples from the RGBNT201 dataset to demonstrate the effectiveness of our caption generation pipeline.
The generated text descriptions provide rich and informative attributes for each spectrum, including clothing, hairstyle and belongings.
These examples verify the effectiveness of our multi-modal captioning pipeline in producing detailed descriptions, which are crucial for multi-modal person ReID under complex visual environments.
\\
\textbf{Multi-modal Vehicle ReID Dataset Examples.}
In \textcolor{red}{Fig.}~\ref{fig:more_instance_vehicle}, we present additional examples from the MSVR310 dataset to illustrate the effectiveness of our caption generation pipeline in the vehicle ReID scenario.
The generated text descriptions capture detailed attributes for each spectrum, including vehicle type, color and license plate number.
Notably, descriptions for RGB images reliably provide accurate license plate numbers, while NIR and TIR images focus on attributes such as vehicle type and logo.
With these detailed and informative text descriptions, we can fully leverage the semantic information from text descriptions to enhance the performance of multi-modal vehicle ReID.
\section{Details of Modules and Experiments}
\subsection{Details of the Proposed Modules in IDEA}
\textbf{Details of Modal Prefixes.}
To differentiate textual descriptions from various modalities while fine-tuning the CLIP text encoder, we introduce Modal Prefixes.
These prefixes consist of two components: a fixed textual description highlighting the characteristics of a specific spectrum and learnable tokens for fine-tuning.
This combination enables modality-aware embedding representations.
To be specific, the Modal Prefixes for each spectrum are defined as follows:
\begin{itemize}
    \item RGB:
    {\small
    \textit{“An image of a XXXX person in the visible spectrum, capturing natural colors and fine details: "}}
    \item NIR:
    {\small
    \textit{“An image of a XXXX person in the near infrared spectrum, capturing contrasts and surface reflectance: "}}
    \item TIR:
    {\small
    \textit{“An image of a XXXX person in the thermal infrared spectrum, capturing heat emissions as temperature gradients: "}}
\end{itemize}
After tokenizing the input text and converting them into embeddings, \textbf{we add randomly initialized learnable tokens} to the embedding at the position corresponding to \textit{“XXXX”}.
During fine-tuning, these learnable tokens are updated to capture semantic information among the text descriptions, providing more information for the subsequent InverseNet.
\\
\textbf{Details of InverseNet.}
To fully leverage the semantic information from text descriptions, we propose the InverseNet.
Specifically, InverseNet $\mathcal{I}$ is a simple MLP layer that takes the text feature $\hat{f}^{t}_{m}$ as input and outputs a pseudo token ${f}^{t}_{m}$.
The input text feature $\hat{f}^{t}_{m}$ is determined based on the presence of learnable tokens in the Modal Prefixes:
\begin{itemize}
    \item If no learnable tokens are added, we directly use the global feature ${f}^{\text{end}}$ from the \textit{“endoftext"} index position of the text, following previous works:
    \begin{equation}
    \hat{f}^{t}_{m} = {f}^{\text{end}}.
    \end{equation}
    \item If learnable tokens are present, we extract the feature corresponding to the position of \textit{“XXXX”}, denoted as ${f}^{\text{prompt}} \in \mathbb{R}^{N_p \times C}$ and average them with ${f}^{\text{end}}$ as follows:
    \begin{equation}
    \hat{f}^{t}_{m} = \frac{1}{N_p + 1} \left( {f}^{\text{end}} + \sum_{i=1}^{N_p} {f}^{\text{prompt}}_i \right).
    \end{equation}
\end{itemize}
This approach ensures that the text representation incorporates both the global context from the \textit{“endoftext"} index and the other semantic information provided by the learnable tokens.
Then, the pseudo token ${f}^{t}_{m}$ is calculated as follows:
\begin{equation}
    {f}^{t}_{m} = \mathcal{I}(\hat{f}^{t}_{m}) = \omega (\varphi (\omega (\delta (\varphi (\hat{f}^{t}_{m}))))),
\end{equation}
where $\varphi$, $\delta$ and $\omega$ denote the linear layer, GeLU activation function and dropout layer, respectively.
Then, the pseudo token ${f}^{t}_{m}$ is concatenated with the visual features, guiding the visual encoder to focus on the interaction between the semantic information and image details.
\\
\textbf{Details of CDA.}
To further enhance the interaction between global features and discriminative local information, we propose CDA.
In this supplementary material, we clarify the following key points regarding CDA:
\begin{itemize}
    \item \textbf{Handling Sampling Point Overflow.}
    During the process of feature extraction, predicted sampling points \(\hat{P}\) may occasionally fall outside the bounds of the feature map.
    To ensure that all sampled locations remain within the valid range, the coordinates are first normalized to the range \([-1, +1]\).
    Subsequently, any sampling point that exceeds this range is clipped to the closest valid value.
    Mathematically, for the sampling point $(\hat{x}, \hat{y})$, the clipping process is defined with the following equations:
    \begin{equation}
    \hat{x} = \text{clip}(\hat{x}, -1, +1), \quad \hat{y} = \text{clip}(\hat{y}, -1, +1).
    \end{equation}
    This clipping process ensures that all coordinates stay within the valid range and allows for stable sampling, particularly at the boundaries of the feature map.

    \item \textbf{Bilinear Interpolation for Feature Sampling.}
    Once the sampling points are determined, we employ bilinear interpolation to extract features from the original feature map \(\hat{F}_m \in \mathbb{R}^{H \times W \times C}\). For each sampling point \((\hat{x}, \hat{y})\), we identify the four neighboring grid points surrounding it.
    Let \((i, j)\) denote the top-left corner of the grid, and the neighboring grid points are then located at \((i+1, j)\), \((i, j+1)\), and \((i+1, j+1)\).
    The interpolation is based on the horizontal and vertical distances between the sampling point \(\hat{P}\) and these neighboring grid points:
    \begin{equation}
    dx = \hat{x} - i, \quad dy = \hat{y} - j.
    \end{equation}
    Then, we calculate the bilinear interpolation weights as:
    \begin{equation}
    w_{mn} = (1 - m \cdot dx)(1 - n \cdot dy), \quad m, n \in \{0, 1\}.
    \end{equation}
    The feature value at the sampled point is then computed as a weighted sum of its neighboring grid points:
    \begin{equation}
      \begin{aligned}
        \hat{F}_m(\hat{x}, \hat{y}) = w_{00}\hat{F}_m(i, j) + w_{10}\hat{F}_m(i+1, j) + \\
        w_{01}\hat{F}_m(i, j+1) + w_{11}\hat{F}_m(i+1, j+1).
      \end{aligned}
  \end{equation}
    This bilinear interpolation procedure ensures that the sampled features are smooth and accurate, effectively preserving discriminative local information.
Finally, we get the sampled features from different modalities and reshape them into the token format, which is then concatenated to get \( F_{S} \in \mathbb{R}^{3N_{S} \times C} \), where \( N_{S} = H_{S} \times W_{S} \).
\end{itemize}
With the above details of our proposed modules, we can effectively leverage the semantic information from text descriptions and enhance the interaction between global features and discriminative local information, leading to superior performance in multi-modal object ReID tasks.
\subsection{Details of Models in the Ablation Study}
\begin{figure}[t]
  \centering
    \resizebox{0.475\textwidth}{!}
    {
  \includegraphics[width=30\linewidth]{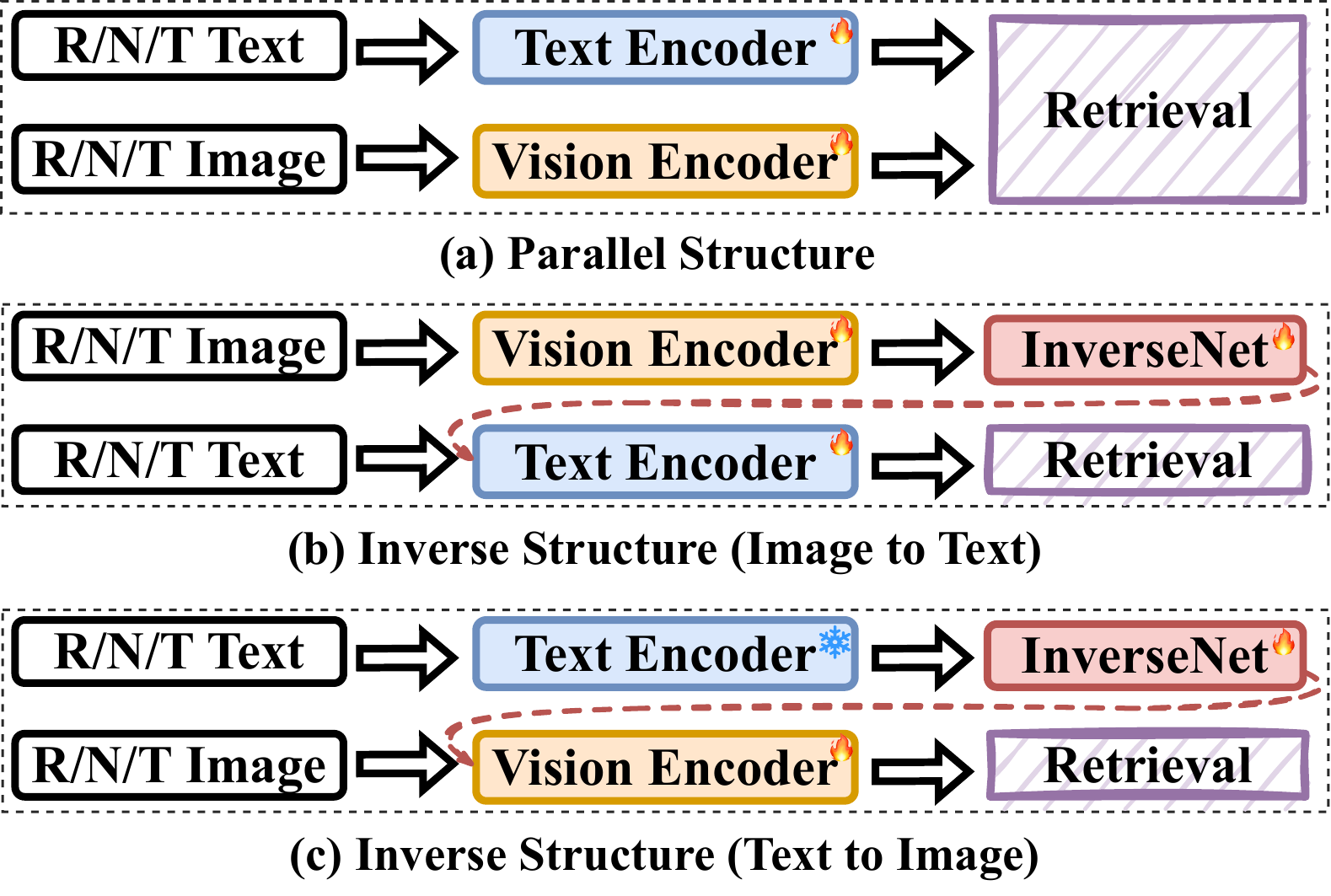}
  }
  \vspace{-4mm}
   \caption{Comparison of different structures in the IMFE.}
  \label{fig:inverse_direction}
  \vspace{-2mm}
\end{figure}

\textbf{Model B in \textcolor{red}{Tab.}~\ref{tab:main_ablation}.}
In \textcolor{red}{Tab.}~\ref{tab:main_ablation} of the main manuscript, Model B integrates text information using a parallel structure.
Specifically, this parallel structure is depicted in \textcolor{red}{Fig.}~\ref{fig:inverse_direction} (a), where the text feature is directly concatenated with the visual feature for retrieval.
In this configuration, the text information is not utilized as effectively as in IMFE.
\\
\textbf{Model A in \textcolor{red}{Tab.}~\ref{tab:CDA_ablation}.}
In \textcolor{red}{Tab.}~\ref{tab:CDA_ablation} of the main manuscript, Model A does not any components in CDA.
All the patch tokens from different modalities are directly concatenated to interact with each other using the self-attention mechanism.
After the interaction, we pool the features from different modalities and concatenate them for retrieval.
Due to the limited local information, the performance of Model A is inferior to the global feature in IMFE.
\\
\textbf{IDEA w/o Text in \textcolor{red}{Tab.}~\ref{tab:Text_ablation}.}
To evaluate the importance of text information in IDEA, we compare the performance of IDEA with and without text information in \textcolor{red}{Tab.}~\ref{tab:Text_ablation}.
In fact, our IDEA can also work without text information.
Under this setting, our IDEA is composed of \textbf{the visual baseline and the CDA}, which only utilizes visual information for retrieval.
The results show that the text information significantly enhances the performance of IDEA, demonstrating the importance of leveraging semantic guidance.
\\
\textbf{IDEA w/o Offset in \textcolor{red}{Tab.}~\ref{tab:Offset_ablation}.}
In \textcolor{red}{Tab.}~\ref{tab:Offset_ablation}, we assess the effectiveness of the offset mechanism in CDA.
When the offset is not used, we rely on convolutional layers to process \( \hat{F}_{m} \), generating features that aggregate local information, which are of the same shape as our \( F_{S} \).
The results demonstrate that the offset mechanism enhances the capture of discriminative local information, leading to improved performance.
\begin{table}[t]
  \centering
  \renewcommand\arraystretch{1.2}
  \setlength\tabcolsep{1.5pt}
  \resizebox{0.478\textwidth}{!}
  {
  \begin{tabular}{ccccccccc}
    \noalign{\hrule height 1pt}
    \multicolumn{1}{c}{\multirow{2}{*}{\textbf{Methods}}} &\multicolumn{1}{c}{\multirow{1}{*}{\textbf{Params}}} &  \multicolumn{2}{c}{\textbf{RGBNT201}} &  \multicolumn{2}{c}{\textbf{RGBNT100}} & \multicolumn{2}{c}{\textbf{MSVR310}} \\
    \cmidrule(r){2-2} \cmidrule(r){3-4} \cmidrule(r){5-6} \cmidrule(r){7-8}
    &\textbf{M} & \textbf{mAP} & \textbf{R-1}& \textbf{mAP} & \textbf{R-1} &\textbf{mAP} & \textbf{R-1} \\
    \hline
  HAMNet~\cite{li2020multi} &  78.00 &27.7 &26.3 &74.5 &93.3 &27.1 &42.3\\
  CCNet~\cite{zheng2023cross} &  74.60 &- &- & 77.2 &96.3 &36.4 &55.2\\
  IEEE~\cite{wang2022interact} & 109.22 &49.5&48.4 &-&-&-&-\\
  GAFNet~\cite{guo2022generative} &  130.00 &- &- &74.4 &93.4 &- &-\\
  \hline
  UniCat$^*$~\cite{crawford2023unicat}  & 259.02 &57.0 &55.7&79.4&96.2&- &- \\
  GraFT$^*$~\cite{yin2023graft} & 101.00 &- &- &76.6&94.3&-&-\\
  TOP-ReID$^*$~\cite{wang2024top} & 324.53 &72.3 &76.6 &81.2&96.4&35.9&44.6\\
  EDITOR$^*$~\cite{zhang2024magic} &118.55 & 66.5       & 68.3& 82.1 & 96.4 &39.0 & 49.3\\
  RSCNet$^*$~\cite{yu2024representation} &  124.10 & 68.2 & 72.5 &82.3 &\underline{96.6} &39.5 &49.6\\
  WTSF-ReID$^*$~\cite{yu2025wtsf} &  143.60 & 67.9 & 72.2 & 82.2 &96.5 &39.2 &49.1\\
  MambaPro$\dagger$~\cite{wang2024mambapro} & 74.20 &78.9&\textbf{83.4}&83.9 &94.7 & \underline{47.0} &56.5\\
  DeMo$\dagger$~\cite{wang2024decoupled} & 98.79 &\underline{79.0}&\underline{82.3}&\underline{86.2} &\textbf{97.6} & \textbf{49.2} &\underline{59.8}\\
  \hline
  \rowcolor[gray]{0.92}
  IDEA$\dagger$ & 91.67 &\textbf{80.2}&82.1&\textbf{87.2} &96.5 & \underline{47.0} &\textbf{62.4}\\
  \noalign{\hrule height 1pt}
  \end{tabular}
  }
  \vspace{-2mm}
  \caption{Parameter and performance comparison with state-of-the-art methods.
  The best and second results are in bold and underlined, respectively.
  The symbol $\dagger$ denotes CLIP-based methods, $*$ indicates ViT-based methods and others are CNN-based methods.}
  \label{tab:params}
  \vspace{-2mm}
  \end{table}
%~~~~~~~~~~~~~~~~~~~~~~~~~~~~~~~~~~~~~~~~~~~~~~~~~~~~~~~~~~~~~~~~~~~~~~~~~~~~~~~~~~~~~~~~~~~~~~~
\section{Module Validation and Analysis}
\begin{figure}[t]
  \centering
    \resizebox{0.475\textwidth}{!}
    {
  \includegraphics[width=30\linewidth]{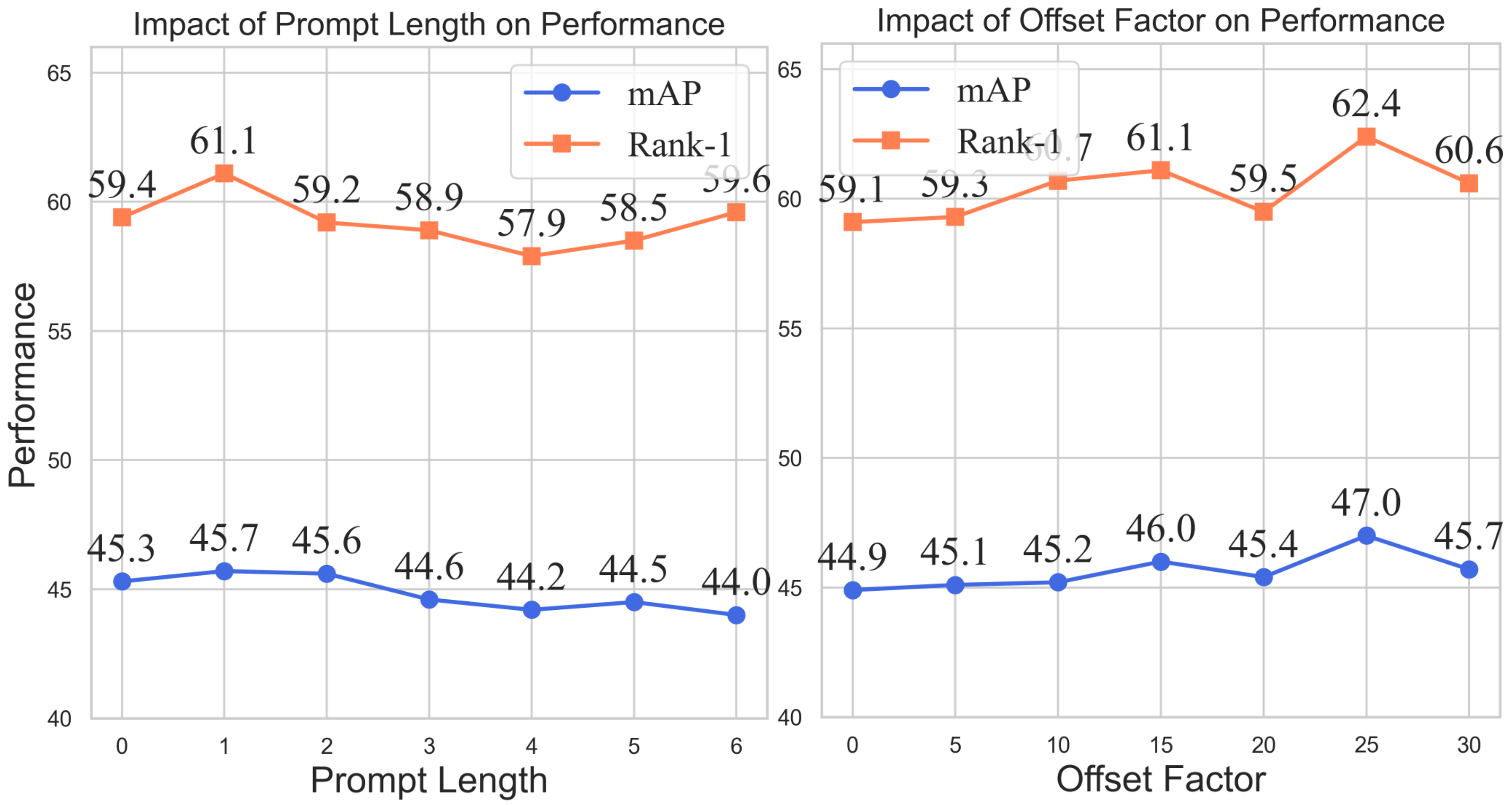}
  }
  \vspace{-3mm}
   \caption{Performance comparison with different prompt lengths and offset factors on multi-modal vehicle dataset MSVR310.}
  \label{fig:prompt_offset_msvr310}
  \vspace{-6mm}
\end{figure}

\textbf{Analysis of Model Parameters and Performance.}
In \textcolor{red}{Tab.}~\ref{tab:params}, we compare the trainable parameters and performance of our proposed IDEA model with several state-of-the-art methods, including CNN-based approaches and Transformer-based techniques.
While Transformer-based methods typically have more parameters than CNN-based approaches, they often deliver superior performance due to their strong generalization capabilities.
Among these, our proposed IDEA stands out by achieving state-of-the-art performance with significantly fewer parameters.
For instance, IDEA requires only 91.67M parameters, far less than TOP-ReID’s 324.53M, yet it achieves substantial improvements in mAP and Rank-1 accuracy.
On the RGBNT201 dataset, IDEA achieves an mAP of 80.2\%, surpassing the previous best of 79.0\% mAP by DeMo.
Similarly, on the RGBNT100 dataset, IDEA achieves an mAP of 87.2\%, which is 1\% higher than DeMo.
These results demonstrate the effectiveness and efficiency of our proposed IDEA.
\begin{table}[t]
  \centering
  \renewcommand\arraystretch{1.1}
  \setlength\tabcolsep{4.5pt}
  \resizebox{0.475\textwidth}{!}
  {
  \begin{tabular}{cccccc}
      \noalign{\hrule height 1pt}
      \multicolumn{1}{c}{\multirow{2}{*}{\textbf{Index}}} &\multirow{2}{*}{\textbf{Inverse Direction}} & \multicolumn{4}{c}{\textbf{Metrics}} \\
      \cmidrule(r){3-6}
          &   & \textbf{mAP}    & \textbf{Rank-1}   & \textbf{Rank-5} & \textbf{Rank-10} \\\hline
  \multirow{1}{*}{A} & Image to Text   &\underline{72.0}  &\underline{73.4} &\underline{83.4}  &\underline{89.2}\\
  \rowcolor[gray]{0.92}
  \multirow{1}{*}{B} & Text to Image    &\textbf{77.2} &\textbf{81.1}  &\textbf{88.4} &\textbf{92.2}\\
  \noalign{\hrule height 1pt}
  \end{tabular}
  }
  \vspace{-2mm}
  \caption{Comparison of inverse directions on RGBNT201.}
  \label{tab:image_inverse}
  \vspace{-2mm}
\end{table}
%~~~~~~~~~~~~~~~~~~~~~~~~~~~~~~~~~~~~~~~~~~~~~~~~~~~~~~~~~~~~~~~~~~~~~~~~~~~~~~~~~~~~~~~~~~~~~~~
\\
\textbf{Comparison with Different Inverse Directions.}
To further validate the effectiveness of our proposed IMFE, we compare the performance of IDEA using different inverse directions on the RGBNT201 dataset.
As shown in \textcolor{red}{Tab.}~\ref{tab:image_inverse}, Model B in \textcolor{red}{Fig.}~\ref{fig:inverse_direction} (c) outperforms Model A in \textcolor{red}{Fig.}~\ref{fig:inverse_direction} (b), achieving an mAP of 77.2\% and Rank-1 accuracy of 81.1\%, compared to 72.0\% mAP and 73.4\% Rank-1 accuracy for the image-to-text direction.
These findings highlight the importance of leveraging semantic guidance from text descriptions to enrich visual interactions.
However, the inherent limitations of multi-modal captioning, where the text modality serves as the primary fusion source, may introduce inconsistencies across spectra, resulting in a performance drop in certain cases.
Thus, we adopt the text-to-image direction in our IDEA model as default setting.
\\
\textbf{Comparison with CLIP-based TOP-ReID.}
In \textcolor{red}{Tab.}~\ref{tab:pre}, we compare our IDEA model with TOP-ReID~\cite{wang2024top}, both utilizing the CLIP vision encoder.
IDEA consistently outperforms TOP-ReID on the RGBNT201 dataset, achieving an mAP of 80.2\% and Rank-1 accuracy of 82.1\%, compared to TOP-ReID’s 73.3\% mAP and 77.2\% Rank-1 accuracy.
These results highlight the effectiveness of our proposed modules in fully leveraging CLIP's knowledge.
\begin{table}[t]
  \vspace{-0mm}
  \centering
  \renewcommand\arraystretch{1.1}
  \setlength\tabcolsep{4.5pt}
  \resizebox{0.475\textwidth}{!}
  {
  \begin{tabular}{cccccc}
    \noalign{\hrule height 1pt}
  {\multirow{2}{*}{\textbf{Methods}}} &{\multirow{1}{*}{\textbf{Params}}}&  \multicolumn{4}{c}{\textbf{Metrics}} \\
  \cmidrule(r){2-2} \cmidrule(r){3-6}
  &\textbf{M} & \textbf{mAP} & \textbf{Rank-1} & \textbf{Rank-5} & \textbf{Rank-10} \\\hline
    TOP-ReID (CLIP) & \multirow{1}{*}{324.53} &\underline{73.3} &\underline{77.2} &\underline{85.9} &\underline{90.1}  \\
    \rowcolor[gray]{0.92}
    IDEA &\multirow{1}{*}{91.67}    &\textbf{80.2} &\textbf{82.1} &\textbf{90.0} &\textbf{93.3}  \\
  \noalign{\hrule height 1pt}
  \end{tabular}
  }
  \vspace{-2mm}
  \caption{Comparison with TOP-ReID (CLIP) on RGBNT201.}
  \vspace{-2mm}
  \label{tab:pre}
\end{table}
\begin{table}[t]
  \centering
  \renewcommand\arraystretch{1.1}
 \setlength\tabcolsep{4.5pt}
  \resizebox{0.475\textwidth}{!}
  { % 
      \begin{tabular}{ccccc}
        \noalign{\hrule height 1pt}
      \textbf{Model} & \textbf{Training Time (h)} & \textbf{Memory (GB)} & \textbf{Time per Epoch (min)} & \textbf{Samples/s} \\
      \midrule
      TOP-ReID     & 0.6650              & 17.80            & 0.3325          & 168.03           \\
      EDITOR       & 0.4074                 & 16.41               & 0.3492              & 167.01             \\
      \rowcolor[gray]{0.92}
      \textbf{IDEA}         & \textbf{0.3075}             & \textbf{18.02}            & \textbf{0.3600}           & \textbf{159.68}           \\
      \noalign{\hrule height 1pt}
      \end{tabular}
  }
  \vspace{-2mm}
  \caption{Training efficiency comparison on RGBNT201.}
  \vspace{-2mm}
  \label{tab:efficiency}
\end{table}
%~~~~~~~~~~~~~~~~~~~~~~~~~~~~~~~~~~~~~~~~~~~~~~~~~~~~~~~~~~~~~~~~~~~~~~~~~~~~~~~~~~~~~~~~~~~~~~~
\\
\textbf{Training Efficiency Comparison.}
As shown in \textcolor{red}{Tab.}~\ref{tab:efficiency}, IDEA demonstrates competitive training efficiency.
Moreover, it requires \textbf{less time to converge}, benefiting from the effective utilization of semantic information from text descriptions.
This guidance enables the model to focus on discriminative local features, accelerating convergence.
Furthermore, GPU memory usage remains \textbf{below 24GB} during both training and inference.
On the smaller RGBNT201 dataset, training can be completed \textbf{within 20 minutes}, further highlighting the efficiency of our proposed IDEA.
\\
\textbf{Effect of Text with Different Modalities Combination.}
In \textcolor{red}{Tab.}~\ref{tab:textIR}, we compare the performance of different modality combinations on the RGBNT201 dataset.
Here, \textbf{R} refers to the RGB image modality along with text annotations derived from RGB images, while \textbf{N} and \textbf{T} represent the NIR and TIR image-text pairs, respectively.
A crucial aspect of multi-modal object ReID is effectively integrating complementary information across modalities.
Even with the addition of textual annotations, multi-spectral information remains indispensable.
As shown in \textcolor{red}{Tab.}~\ref{tab:textIR}, incorporating IR data significantly enhances performance, underscoring its critical role in capturing essential identity cues.
The results highlight that while text provides valuable semantic guidance, it cannot fully replace the rich visual and spectral details contributed by multi-modal fusion.
\begin{table}[t]
    \centering
    \renewcommand\arraystretch{1.1}
   \setlength\tabcolsep{4.5pt}
    \resizebox{0.475\textwidth}{!}
    { % 
        \begin{tabular}{cccccccc}
          \noalign{\hrule height 1pt}
        \textbf{Modality} &\textbf{R} & \textbf{N} & \textbf{T} & \textbf{R+N} & \textbf{R+T}& \textbf{N+T}& \textbf{R+N+T} \\
        \midrule
        mAP &39.9      & 27.1       &43.3& 58.4 & 71.5         & 62.9               & 80.2                     \\
        \noalign{\hrule height 1pt}
        \end{tabular}
    }
    \caption{\small Comparison of modalities combination on RGBNT201.}
    \label{tab:textIR}
\end{table}
\\
%~~~~~~~~~~~~~~~~~~~~~~~~~~~~~~~~~~~~~~~~~~~~~~~~~~~~~~~~~~~~~~~~~~~~~~~~~~~~~~~~~~~~~~~~~~~~~~~
\textbf{Exploration of Extreme Cases with Missing Text.}
In \textcolor{red}{Tab.}~\ref{tab:textMissing}, we explore the impact of missing text annotations during retrieval on the RGBNT201 dataset.
Notably, in the most extreme case where all text annotations are removed, IDEA still achieves a mAP of 79.5\%, which is only a marginal drop from the fully annotated setting.
This indicates that during training, the semantic guidance provided by text effectively enhances the fusion between multi-modal features, enabling the model to learn more robust representations.
These findings validate that IDEA does not excessively rely on textual input during inference but instead leverages structured multi-modal learning to develop a more robust and adaptive feature representation.
\begin{table}[t]
  \centering
  \renewcommand\arraystretch{1.1}
   \setlength\tabcolsep{4.5pt}
  \resizebox{0.475\textwidth}{!}
  { % 
      \begin{tabular}{ccccccccc}
        \noalign{\hrule height 1pt}
      \textbf{Cases} &\textbf{Full} & \textbf{M(R)} & \textbf{M(N)} & \textbf{M(T)} & \textbf{M(RN)}& \textbf{M(RT)}& \textbf{M(NT)}& \textbf{M(RNT)} \\
      \midrule
      mAP &80.2       & 79.8       &80.0& 79.9 & 79.7         & 79.6               & 79.8              & 79.5             \\
      \noalign{\hrule height 1pt}
      \end{tabular}
  }
  \caption{Comparison of text-missing cases on RGBNT201.
  M(X) indicates that the text annotations for modality X are blank strings.}
  \label{tab:textMissing}
\end{table}
\begin{table}[t]
  \centering
  \renewcommand\arraystretch{1.}
  \setlength\tabcolsep{4.5pt}
  \resizebox{0.35\textwidth}{!}
  {
  \begin{tabular}{cccccc}
      \noalign{\hrule height 1pt}
      \multicolumn{1}{c}{\multirow{2}{*}{\textbf{Index}}} &\multicolumn{3}{c}{\textbf{Modules}} & \multicolumn{2}{c}{\textbf{Metrics}} \\
      \cmidrule(r){2-4} \cmidrule(r){5-6}
 & \textbf{Text}              & \textbf{IMFE}                & \textbf{CDA}                   & \textbf{mAP}    & \textbf{Rank-1}   \\\hline
  A                  & \ding{53}                  & \ding{53}                  & \ding{53}                    & 40.4  & 56.0 \\
  B                  & \ding{51}                  & \ding{53}                  & \ding{53}                      & 43.5  & 57.9 \\
  \multirow{1}{*}{C} & \multirow{1}{*}{\ding{51}} & \multirow{1}{*}{\ding{51}} & \multirow{1}{*}{\ding{53}}    & \underline{45.7}  & \underline{61.1} \\
  \rowcolor[gray]{0.92}
  \multirow{1}{*}{D} & \multirow{1}{*}{\ding{51}} & \multirow{1}{*}{\ding{51}} & \multirow{1}{*}{\ding{51}}    &\textbf{47.0} &\textbf{62.4}  \\
  \noalign{\hrule height 1pt}
  \end{tabular}
  }
  \vspace{-2mm}
  \caption{Comparison with different modules on MSVR310.}
  \label{tab:main_ablation_vehicle}
  \vspace{-6mm}
\end{table}
%~~~~~~~~~~~~~~~~~~~~~~~~~~~~~~~~~~~~~~~~~~~~~~~~~~~~~~~~~~~~~~~~~~~~~~~~~~~~~~~~~~~~~~~~~~~~~~~
\\
\textbf{Effect of Key Modules on Vehicle Dataset.}
In \textcolor{red}{Tab.}~\ref{tab:main_ablation_vehicle}, we present the performance comparison of different modules on the MSVR310 vehicle dataset.
Model A, which only uses visual information, achieves an mAP of 40.4\% and Rank-1 accuracy of 56.0\%.
Model B introduces the text information with a parallel structure, leading to a 3.1\% improvement in mAP compared to Model A.
Model C further incorporates the IMFE, which fully leverages the semantic guidance from the text, resulting in a 2.2\% improvement in mAP.
Finally, Model D integrates the CDA, leading to a 1.3\% improvement in mAP compared to Model C, achieving the best performance with an mAP of 47.0\% and Rank-1 accuracy of 62.4\%.
These results fully validate the effectiveness and efficiency of our proposed modules.
\\
\textbf{Hyper-parameter Analysis for Vehicle Dataset.}
\textcolor{red}{Fig.}~\ref{fig:prompt_offset_msvr310} presents a performance comparison across different prompt lengths and offset factors on the MSVR310 vehicle dataset.
The results demonstrate a general performance improvement with increasing prompt length.
However, performance begins to decline when the prompt length exceeds 1.
Consequently, we adopt a prompt length of 1 for subsequent experiments on MSVR310.
A similar observation holds for the RGBNT201 dataset, where longer prompts truncate text descriptions, leading to the loss of critical information.
For the offset factor, the performance remains relatively stable, with the optimal result achieved at an offset factor of 25.
%～～～～～～～～～～～～～～～～～～～～～～～～～～～～～～～～
\section{Visualization Analysis of IDEA}
\subsection{More Visualization for Person ReID}
\begin{figure}[t]
  \centering
    \resizebox{0.475\textwidth}{!}
    {
  \includegraphics[width=30.\linewidth]{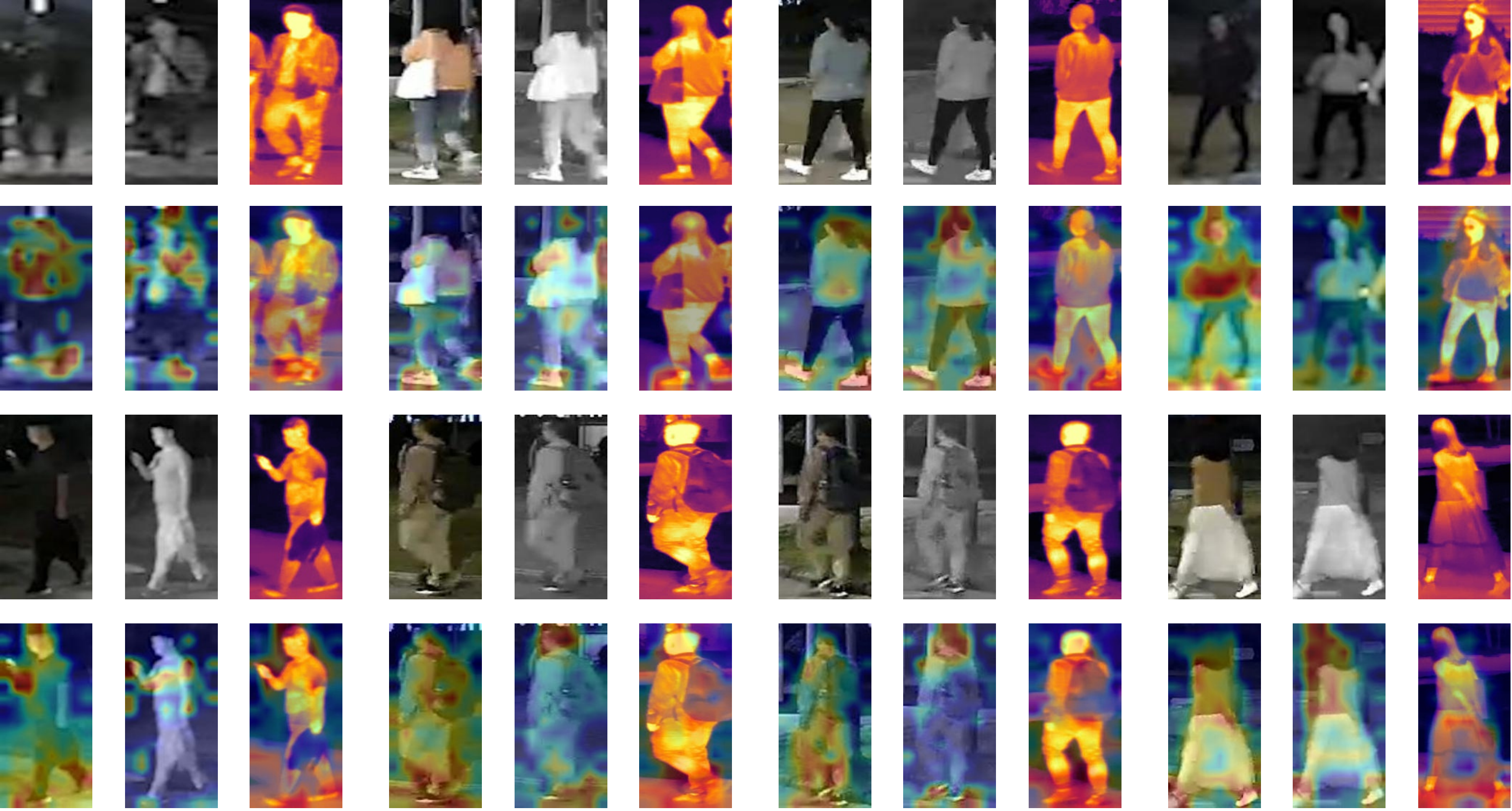}
  }
  \vspace{-6mm}
   \caption{Visualization of channel activation maps of different modalities on the person ReID dataset.}
  \label{fig:channel_act}
  \vspace{-4mm}
\end{figure}

\begin{figure*}[t]
  \centering
    \resizebox{0.96\textwidth}{!}
    {
  \includegraphics[width=30.\linewidth]{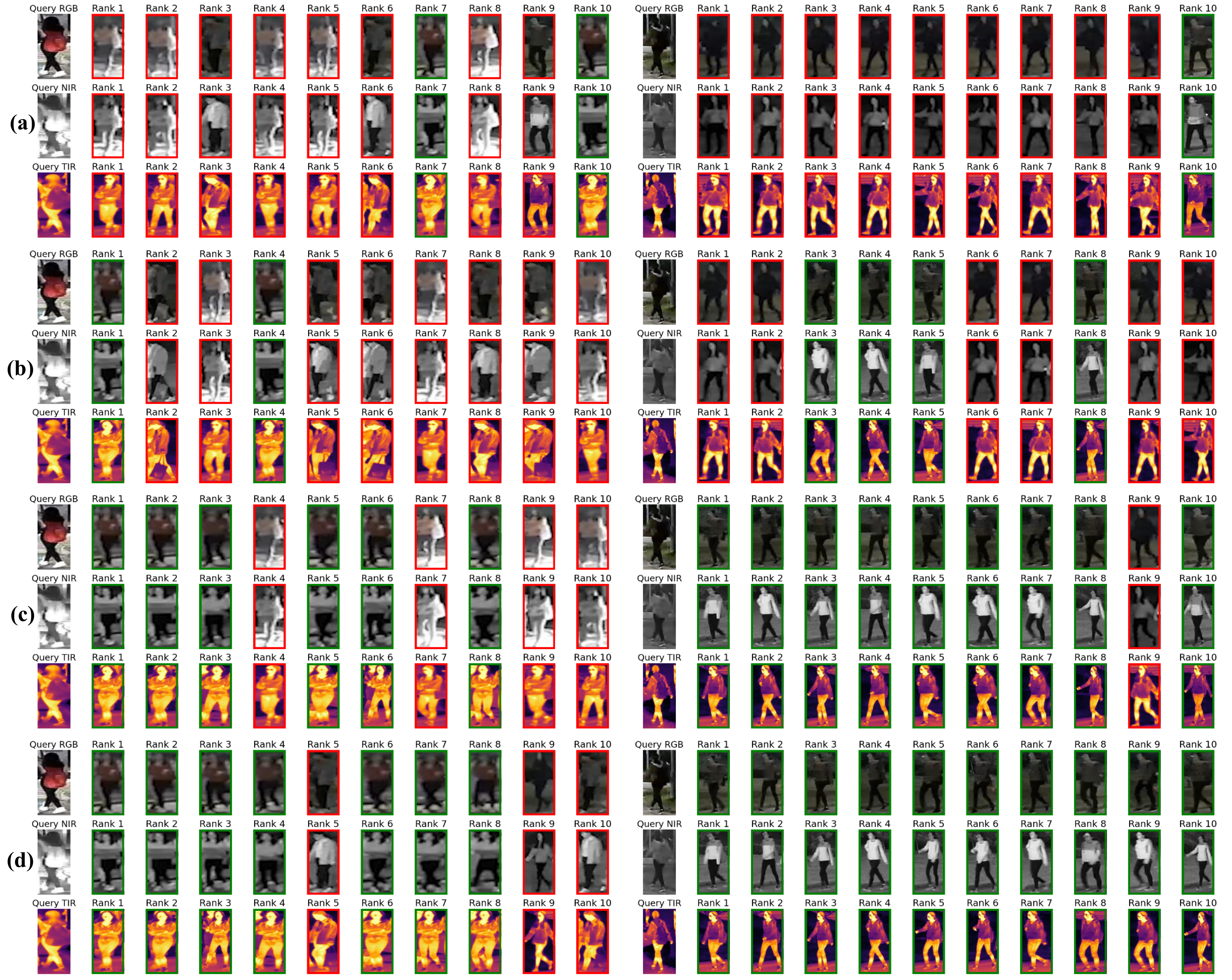}
  }
  \vspace{-2mm}
   \caption{Ranking list comparison with different modules on the person ReID dataset RGBNT201.
   (a) Baseline.
   (b) Baseline + Text.
   (c) Baseline + IMFE.
   (d) Baseline + IMFE + CDA.
   The green box indicates the correct match, while the red box indicates the incorrect match.}
  \label{fig:rank_overall}
  \vspace{-2mm}
\end{figure*}

\begin{figure*}[t]
  \centering
    \resizebox{0.96\textwidth}{!}
    {
  \includegraphics[width=30.\linewidth]{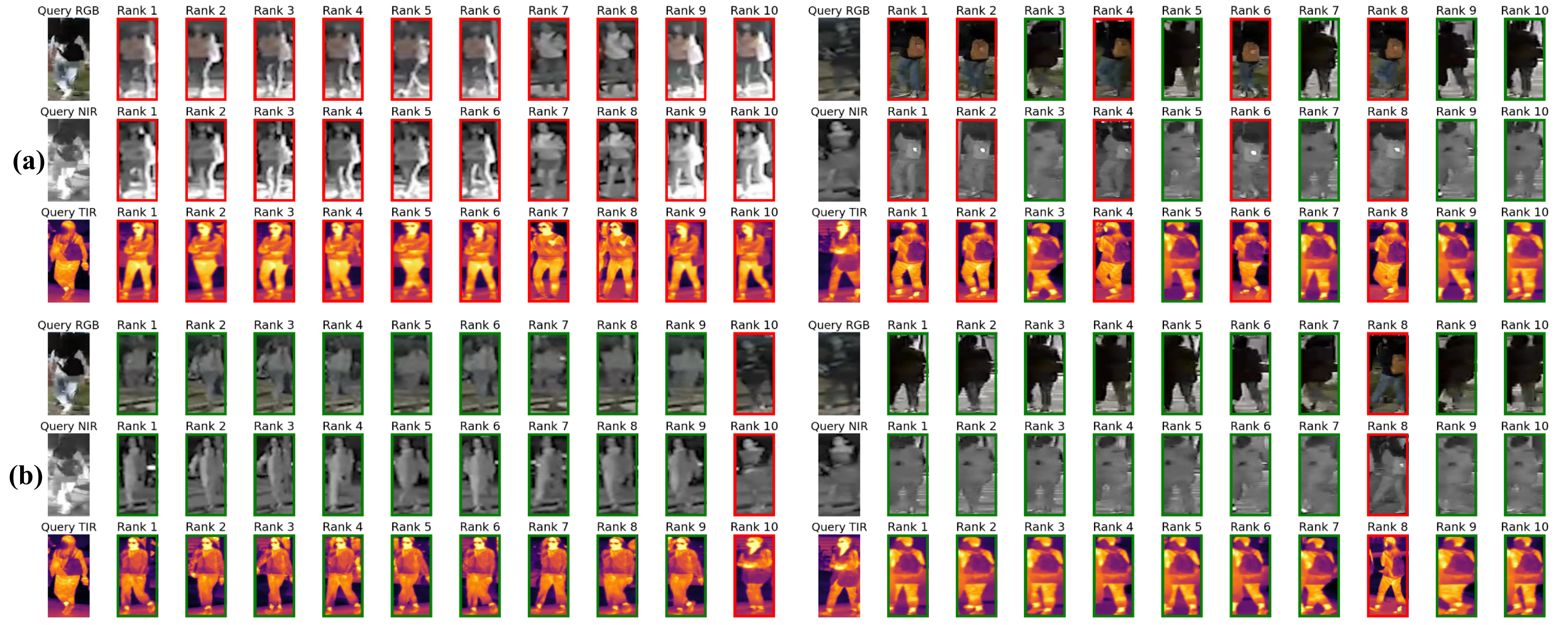}
  }
  \vspace{-2mm}
   \caption{Multi-modal ranking list comparison with TOP-ReID on the person ReID dataset RGBNT201.
   (a) TOP-ReID.
    (b) IDEA.
    The multi-modal ranking list of TOP-ReID is sourced from the supplementary material of DeMo~\cite{wang2024decoupled}.}
  \label{fig:rank_TOP}
  \vspace{-2mm}
\end{figure*}

\textbf{Visualization of Channel Activation Maps on the Person ReID Dataset.}
\textcolor{red}{Fig.}~\ref{fig:channel_act} illustrates the channel activation maps across different modalities for person ReID datasets within our IDEA framework.
Each modality exhibits distinct activation patterns, reflecting their unique spectral properties.
Notably, these maps successfully identify semantic regions, including hair, clothing and accessories, underscoring the capability of our proposed modules to effectively leverage multi-modal information.
\\
\textbf{Visualization of Multi-modal Ranking List with Different Modules in IDEA.}
%~~~~~~~~~~~~~~~~~~~~~~~~~~~~~~~~~~~~~~~~~~~~~~~~~~~~~~~~~~~~~~~~~~~~~~~~~~~~~~~~~~~~~~~~~~~~~~~
\begin{figure}[t]
  \centering
    \resizebox{0.475\textwidth}{!}
    {
  \includegraphics[width=30.\linewidth]{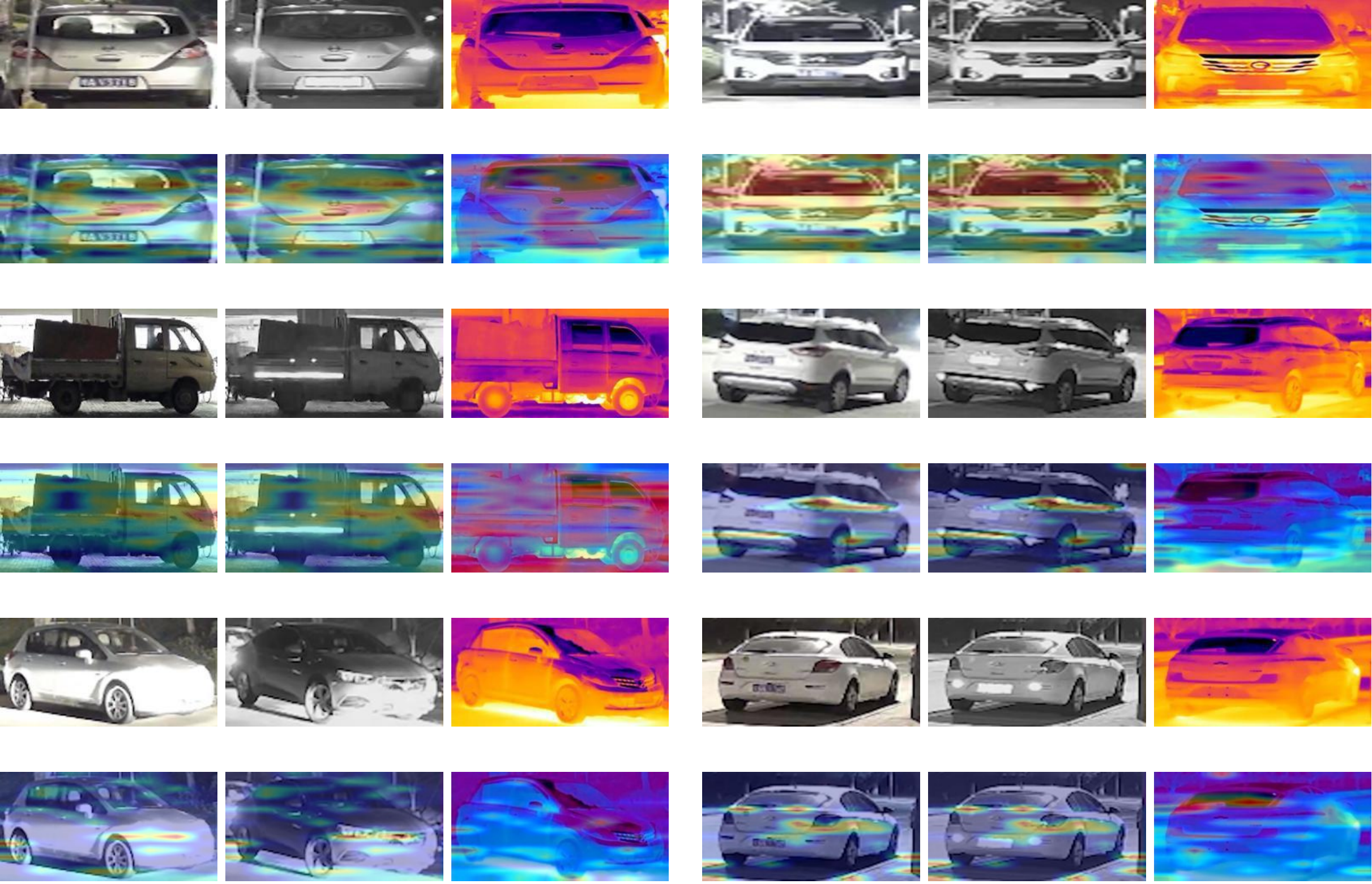}
  }
  \vspace{-6mm}
   \caption{Visualization of channel activation maps of different modalities on the vehicle ReID dataset.}
  \label{fig:channel_act_vehicle}
  \vspace{-4mm}
\end{figure}
%~~~~~~~~~~~~~~~~~~~~~~~~~~~~~~~~~~~~~~~~~~~~~~~~~~~~~~~~~~~~~~~~~~~~~~~~~~~~~~~~~~~~~~~~~~~~~~~
In \textcolor{red}{Fig.}~\ref{fig:rank_overall}, we present the multi-modal ranking list comparison with different modules on the RGBNT201 dataset.
The baseline model without text information exhibits a significant number of incorrect matches, as shown in \textcolor{red}{Fig.}~\ref{fig:rank_overall} (a).
By incorporating text information, the model can better distinguish between correct and incorrect matches, as shown in \textcolor{red}{Fig.}~\ref{fig:rank_overall} (b).
The introduction of IMFE further enhances the model's ability to identify correct matches, as shown in \textcolor{red}{Fig.}~\ref{fig:rank_overall} (c).
Finally, the integration of CDA significantly improves the model's performance, as shown in \textcolor{red}{Fig.}~\ref{fig:rank_overall} (d).
These results demonstrate the effectiveness of our proposed modules in enhancing the performance of multi-modal object ReID.
%~~~~~~~~~~~~~~~~~~~~~~~~~~~~~~~~~~~~~~~~~~~~~~~~~~~~~~~~~~~~~~~~~~~~~~~~~~~~~~~~~~~~~~~~~~~~~~~
\\
\textbf{Visualization of Multi-modal Ranking List Comparison with TOP-ReID.}
In \textcolor{red}{Fig.}~\ref{fig:rank_TOP}, we present the multi-modal ranking list comparison with TOP-ReID on the RGBNT201 person dataset.
We choose hard examples that TOP-ReID struggles with and compare the performance of our IDEA model.
The results verify that our IDEA model consistently outperforms TOP-ReID, achieving accurate matches.
%～～～～～～～～～～～～～～～～～～～～～～～～～～～～～～～～
%～～～～～～～～～～～～～～～～～～～～～～～～～～～～～～～～
\begin{figure*}[t]
  \centering
    \resizebox{0.94\textwidth}{!}
    {
  \includegraphics[width=30.\linewidth]{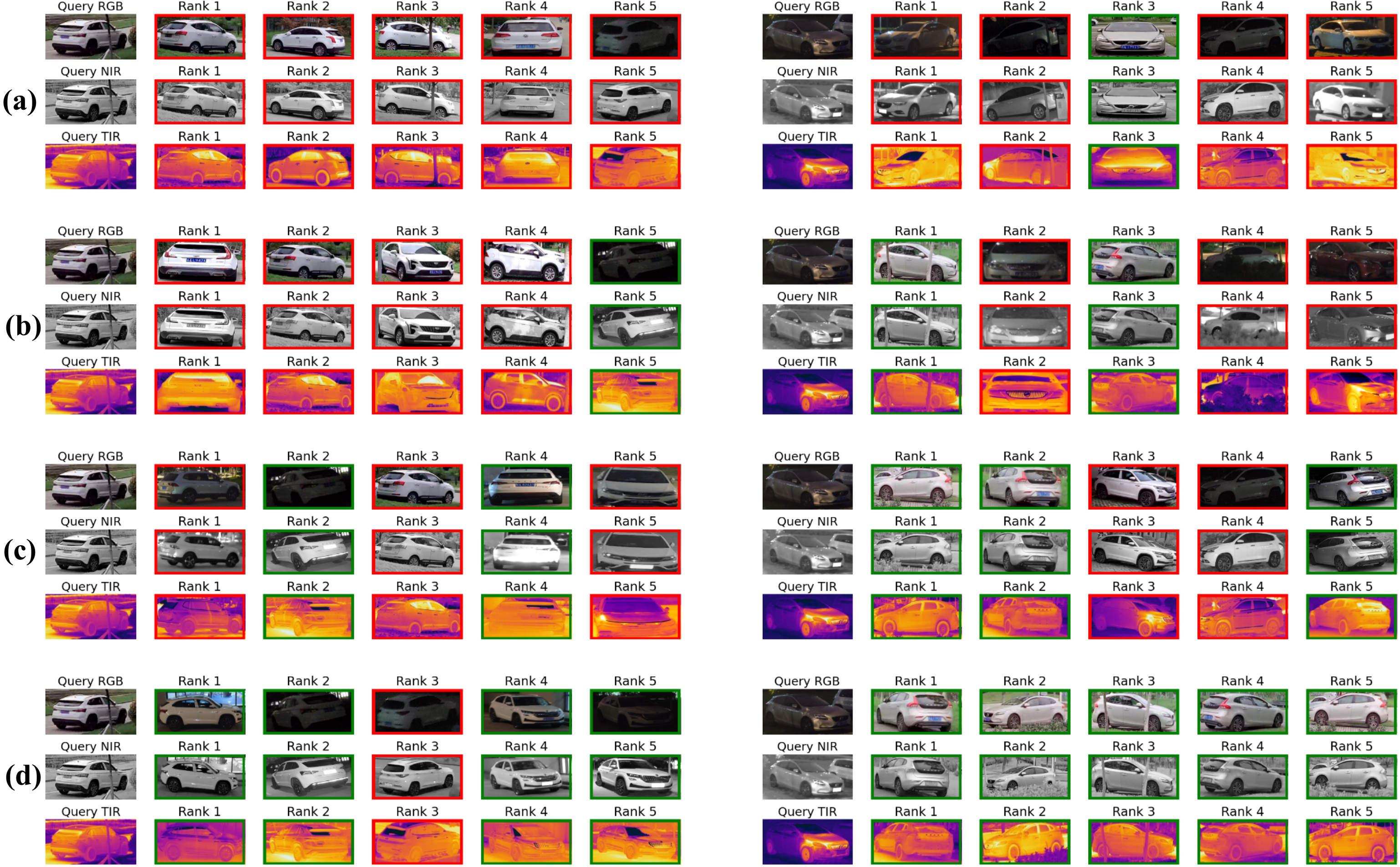}
  }
  \vspace{-2mm}
   \caption{Ranking list comparison with different modules on the vehicle ReID dataset MSVR310.
   (a) Baseline.
   (b) Baseline + Text.
   (c) Baseline + IMFE.
   (d) Baseline + IMFE + CDA.}
  \label{fig:rank_vehicle_modules}
  \vspace{-2mm}
\end{figure*}
%～～～～～～～～～～～～～～～～～～～～～～～～～～～～～～～～
\begin{figure*}[t]
  \centering
    \resizebox{0.94\textwidth}{!}
    {
  \includegraphics[width=30.\linewidth]{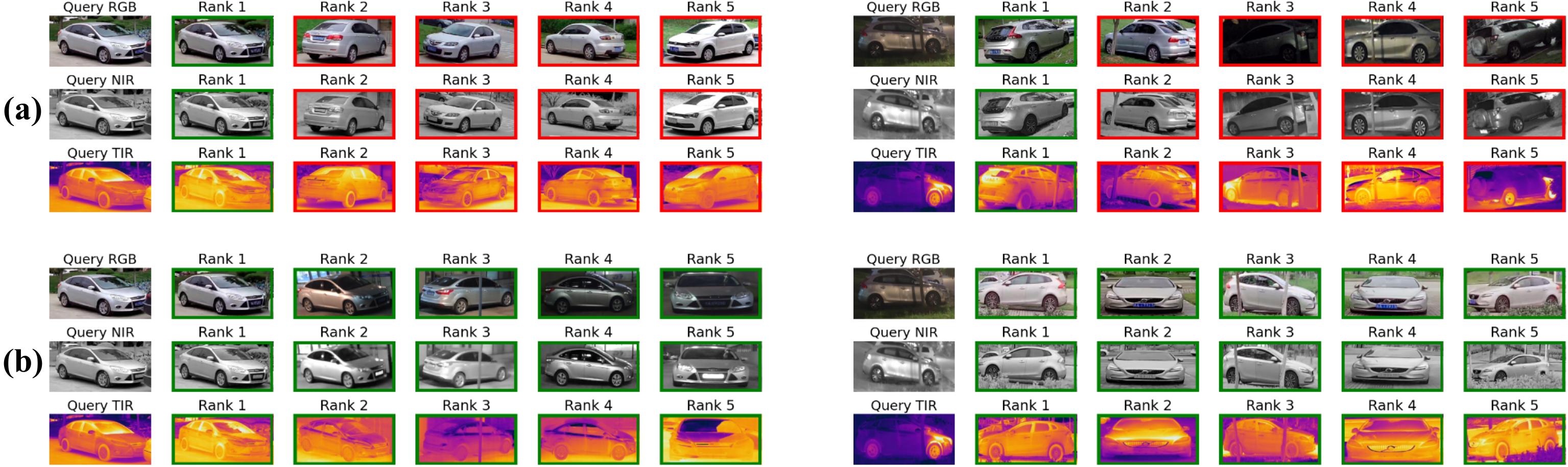}
  }
  \vspace{-2mm}
   \caption{Multi-modal ranking list comparison with EDITOR on the vehicle ReID dataset MSVR310.
   (a) EDITOR.
    (b) IDEA.}
  \label{fig:rank_EDITOR}
  \vspace{-2mm}
\end{figure*}
%～～～～～～～～～～～～～～～～～～～～～～～～～～～～～～～～
\subsection{More Visualization for Vehicle ReID}
\textbf{Visualization of Channel Activation Maps on the Vehicle ReID Dataset.}
In \textcolor{red}{Fig.}~\ref{fig:channel_act_vehicle}, we present the channel activation maps of different modalities in IDEA on the vehicle ReID dataset.
Different modalities focus on distinct semantic regions, such as the corners of the vehicle and the license plate.
These activation maps effectively capture discriminative local information, highlighting the importance of leveraging interaction between global features and discriminative local information in multi-modal vehicle ReID.
\\
\textbf{Visualization of Multi-modal Ranking List with Different Modules.}
\textcolor{red}{Fig.}~\ref{fig:rank_vehicle_modules} provides a comparative analysis of multi-modal ranking lists across different module configurations on the MSVR310 vehicle dataset.
The baseline model, as seen in \textcolor{red}{Fig.}~\ref{fig:rank_vehicle_modules} (a), relies solely on visual features and fails to reliably distinguish between vehicles with similar appearances, leading to frequent mismatches.
Integrating textual information into the model, as shown in \textcolor{red}{Fig.}~\ref{fig:rank_vehicle_modules} (b), greatly enhances the retrieval process, enabling the system to utilize semantic cues for better discrimination.
With the introduction of IMFE (\textcolor{red}{Fig.}~\ref{fig:rank_vehicle_modules} (c)), the model demonstrates an ability to capture more nuanced modality-specific information, resulting in a marked improvement in retrieval accuracy.
Finally, the addition of CDA (\textcolor{red}{Fig.}~\ref{fig:rank_vehicle_modules} (d)) further refines the ranking results, achieving a level of performance that effectively balances semantic understanding and visual precision.
These findings validate the effectiveness of our modules in addressing the challenges of vehicle ReID.
\\
\textbf{Visualization of Multi-modal Ranking List Comparison with EDITOR.}
In \textcolor{red}{Fig.}~\ref{fig:rank_EDITOR}, we compare our IDEA model against EDITOR, the previous state-of-the-art approach on the MSVR310 dataset.
EDITOR, while highly effective in many scenarios, struggles with difficult cases involving visually similar vehicles or missing modality-specific details.
This limitation is evident in its inability to consistently rank correct matches, especially when semantic guidance is needed to resolve ambiguities.
In contrast, our IDEA model leverages advanced multi-modal modules to seamlessly integrate visual and textual cues, resulting in a dramatic reduction in errors even for the hard cases.
This comparative analysis highlights IDEA's superiority over EDITOR, verifying its robustness in tackling real-world challenges.

\end{document}